\documentclass[final,3p,times,authoryear]{elsarticle}

\usepackage{amssymb}
\usepackage{multirow}
\usepackage{algorithm}
\usepackage{algpseudocode}
\usepackage{longtable}
\usepackage{arydshln}
\usepackage{bm}
\usepackage[super]{nth}
\usepackage[inline]{enumitem}
\usepackage{subcaption}
\usepackage{url}
\usepackage{amsthm}
\usepackage{amsmath}
\usepackage{amsfonts}
\usepackage{booktabs}
\usepackage{placeins}

\journal{Transportation Research Part C: Emerging Technologies}

\begin{document}

\begin{frontmatter}



\title{RUMBoost: Gradient Boosted Random Utility Models}


\author[inst1]{Nicolas Salvadé\corref{cor1}}
\ead{nicolas.salvade.22@ucl.ac.uk}
\cortext[cor1]{Corresponding author}
\affiliation[inst1]{organization={Department of Civil, Environmental and Geomatic Engineering, University College London},
           addressline={Gower Street}, 
            city={London},
            postcode={WC1E 6BT}, 
            country={United Kingdom}}

\author[inst1]{Tim Hillel}
\ead{tim.hillel@ucl.ac.uk}

\begin{abstract}
This paper introduces the RUMBoost model, a novel discrete 
choice modelling approach that combines the interpretability 
and behavioural robustness of Random Utility Models (RUMs) with the 
generalisation and predictive ability of deep learning methods. We obtain 
the full functional form of non-linear utility specifications by replacing 
each linear parameter in the utility functions of a RUM with an ensemble of gradient boosted 
regression trees. This enables piece-wise constant 
utility values to be imputed for all alternatives directly from the data for any possible combination of input variables. We introduce additional constraints on the 
ensembles to ensure three crucial features of the 
utility specifications: (i) dependency of the utilities of each alternative on only
the attributes of that alternative, (ii) monotonicity of marginal utilities, and 
(iii) an intrinsically interpretable functional form, where the exact response of the 
model is known throughout the entire input space. Furthermore, we introduce an optimisation-based 
smoothing technique that replaces the
piece-wise constant utility values of alternative attributes with 
monotonic piece-wise cubic splines to identify non-linear parameters with
defined gradient. We demonstrate the potential of the RUMBoost model 
compared to various ML and Random Utility benchmark models for revealed preference mode choice data 
from London. The results highlight both the great predictive performance and the 
direct interpretability of our proposed approach. Furthermore, by analysing the non-linear utility functions, we can identify complex behaviours associated with
different transportation modes which would not have been possible 
with conventional approaches. The smoothed attribute utility 
functions allow for the calculation of various behavioural indicators 
such as the Value of Time (VoT) and marginal utilities. Finally, we demonstrate the flexibility of our 
methodology by showing how the RUMBoost model can be
extended to complex model specifications, including attribute interactions, 
correlation within alternative error terms (Nested Logit model) and heterogeneity 
within the population (Mixed Logit model).
\end{abstract}



\begin{keyword}
Discrete Choice \sep Mode Choice \sep Machine Learning \sep Random Utility \sep Ensemble Learning


\end{keyword}

\end{frontmatter}



\section{Introduction and literature review}
\label{Introduction}

Discrete choice models (DCMs), based on Random Utility theory, have been used 
extensively to model choices over the last 50 years
\citep{Ben-Akiva_Lerman_1985, train2009discrete}, including the choice of travel mode. DCMs have many desirable
qualities: most crucially, their parametric form is directly interpretable and
allows for the integration of expert knowledge consistent with behavioural theory. For example, with a DCM, it is possible to: 
\begin{enumerate*}[label=(\roman*)]
    \item ensure marginal utilities in 
    the model are monotonic (e.g. that increasing the cost of an alternative 
    will always decrease its utility);
    \item restrict the utilities of each alternative 
    to be dependent only on the attributes (or features) of that alternative; and 
    \item define arbitrary interactions of socio-economic characteristics and alternative-specific attributes.
\end{enumerate*}    
These three points are 
essential to derive key behavioural indicators, such as elasticities and Value of Time (VoT), used to inform transport policies and investment decisions. 
However, the parametric form of DCMs is not without disadvantages. Crucially, the linear-in-parameters utility functions are relatively inflexible and must be specified in advance by the modeller. 
As such, these models may fail to capture complex phenomena and non-linear effects in human behaviour.
Among DCMs, one of the best-known and most widely used models is 
the Multinomial Logit model (MNL) derived by \citet{mcfadden1973conditional}. The 
model assumes that the error term is a type 1 extreme value (EV) random variable, which is independent and identically distributed 
(\textit{i.i.d.}) across all alternatives and observations. This formulation results in a closed-form expression of the probability, allowing for easier estimation of the model parameters. However, more complex model specifications exist, for example to capture:
\begin{enumerate*}[label=(\roman*)]
    \item different distributions of the error term; 
    \item nesting (joint consideration) of alternatives; 
    \item behavioural heterogeneity across individuals in the population; and
    \item behavioural heterogeneity across sequential choices. 
\end{enumerate*}

There have been numerous attempts to apply machine learning (ML) probabilistic classification algorithms, such as neural networks and ensemble learning, to investigate choice 
behaviour. These models exhibit high predictive performance and, thanks to
their data-driven nature, do not require any utility functions to be specified 
in advance of model estimation. However, they lack an underlying 
behavioural model and so it is not possible to guarantee consistency of forecasts or derive behavioural indicators such as Value of Time (VoT) or willingness-to-
pay from the model parameters. 
Initial approaches for analysing these models from a behavioural perspective rely on approximating the partial derivatives of the output probabilities of unconstrained ML classifiers in order to define elasticities (and therefore marginal rates of substitution\slash VoT) for variables of interest. 
\citet{wang_deep_2020} use this approach to extract economic indicators from neural networks for mode choice problems, whilst \citet{martin-baos_prediction_2023} extend this to several other classification algorithms, including Gradient Boosting Decision Trees, the Random Forest, and Support Vector Machines. 
Unlike marginal utilities from a DCM, which have a parametric functional form, the probability derivatives of ML classifiers provide only a numeric estimate of the point elasticities at observed data points.
Furthermore, as the underlying models are unconstrained, they exhibit several qualities that are inconsistent with random utility theory, including: 
\begin{enumerate*}[label=(\roman*)]
    \item non-monotonic elasticities, leading to  unwanted behaviours, such as a negative VoT; and 
    \item including all features for all alternatives uniformly, therefore violating the independence of irrelevant alternatives assumption. 
\end{enumerate*}
As such, these techniques have seen limited real-world 
use and practitioners continue to rely predominantly on parametric DCMs. That being
said, the ability of ML models to capture complex non-linear relationships
as well as their improved predictive accuracy makes them an attractive
proposition. 

In response to these limitations, there has been an emergence of hybrid \emph{data-driven utility models} in recent years, that attempt to combine the benefits of ML and DCMs. These can largely be grouped into two different approaches:
\begin{enumerate}
    \item adding additional constraints to machine learning models (e.g. monotonicity, alternative-specific attributes, etc) so that their output can mimic DCM utility values; and 
    \item using data-driven approaches to automate or assist with identifying suitable parametric utility functions. 
\end{enumerate}

There have been several studies which attempt to incorporate key components from DCMs through making appropriate constraints on the structure of ML models.
This analysis stems from the fact that probabilistic ML classifiers, such as neural networks and GBDTs, make use of the same logistic function (commonly referred to as \emph{softmax} in ML) as the MNL to generate choice probabilities over each alternative; replacing the linear-in-parameter utility functions of the MNL with a complex network of neurons with non-linear activation functions (in the case of the neural network) or ensemble of regression trees (in the case of GBDTs). 
As such, with suitable constraints on the model, the pre-softmax regression values for each alternative can be considered as \emph{pseudo-utilities}. 
The most important of these constraints is that the pseudo-utility of each alternative is a function of only the attributes of that alternative (alongside the socio-economic characteristics of the decision-maker), therefore obeying the independence of irrelevant alternatives assumption. 
Without this assumption, it is not possible to perform a utility-based behavioural analysis as modifying one attribute would affect the utility values of all alternatives. 
\citet{wang_deep_2021} build on their prior work of modelling choices with DNNs by defining a separate sub-network per 
alternative, dependent only on the attributes of that alternative, thus allowing for more robust behavioural indicators to be extracted. Similarly, \citet{martin2021revisiting} make use of kernel Logistic Regression (KLR) (where the non-linearity is achieved through transforming the input data with a kernel) to
learn the deterministic part of each alternative utility by having one kernel
per alternative. This allows for behavioural indicators to be defined in the pseudo-utility space, rather than the probability space. 
Other studies instead use a neural network to complement a conventional utility model. 
\citet{sifringer_enhancing_2020} use a Convolutional Neural Network 
(CNN) to learn a linear-in-parameter utility function for the alternative-specific attributes alongside a DNN to get socio-economic interactions, adding them to the pseudo-utility output in a second stage. 
In a similar approach, \citet{wong_reslogit_2021} use a Residual Neural Network to add
a non-linear, fully-connected residual on top of the linear pseudo-utility at each 
layer of the network. 

However, these studies still lack of a key component of the interpretability and 
extrapolation ability of DCMs: monotonicity of marginal pseudo-utilities. Whilst not relying on traditional deep learning models, there are several studies that allow for monotonicity of marginal utilities in ML models.
\citet{kim2023new} use a lattice network with input and output calibration layer
to account for non-linearities and partial monotonicity. 
\citet{krueger_stated_2022} replace the sum operator in a traditional Mixed Logit
model with the Choquet integral, accounting for attribute interaction under 
monotonicity. Lastly, \citet{aboutaleb2022theory} take advantage of the sum of 
squares of polynomials to learn data-driven utility functions under monotonicity 
and shape constraints.

Finally, some researchers retain the traditional structure of the utility 
function but add data-driven components such as learned parameters or assisted 
utility specification. \citet{han2022neural}, with the TasteNet model, use a 
Deep Neural Network to learn individual-specific parameters from socio-economic
characteristics, constraining the sign of the parameters with the appropriate
activation function. Instead of replacing the deterministic
utility by a data driven model, \citet{hillel2019weak} and 
\citet{ortelli_assisted_2021} use data-driven models to assist the modeller with
utility specification. The first paper suggests potential non-linear interactions 
of attributes from the gain of split points in a GBDT algorithm. The second study
proposes an algorithm for automatically evaluating multiple model specifications by
drawing the Pareto frontier, which is useful in assessing the quality of model 
specification under multiple objectives. However, the modeller still needs to
pre-specify the input space and identify non-linearities.

The summary of state-of-the-art practices in hybrid utility models is in Table \ref{contribution_summary}.
The features that we consider in the table are if the models:
\begin{enumerate*}[label=(\roman*)]
    \item include alternative-specific attributes;
    \item incorporate monotonicity constraints;
    \item have intrinsically interpretable pseudo-utilities, as in DCMs;
    \item provide automatic identification of non-linearity in the pseudo-utilities, i.e the complete non-linear transformations from the input variable to the pseudo-utilities are observable; and
    \item include non-i.i.d. error terms, therefore allowing for complex model specifications such as the Nested Logit or the Mixed Logit models;
\end{enumerate*}
In this paper, we provide an algorithm to directly learn the non-linear full 
functional form of the utility, while ensuring alternative-specific attributes and
monotonicity on key attributes.

\begin{table}[ht]
\centering
\caption{\centering Summary of state-of-the-art practice in hybrid utility models. A cross (X) means that the model implements the feature and a tilde ($\sim$) means that the model partly incorporates the feature. Complete explanations on the features that we consider for each column are in the text.}
\begin{tabular}{lccccc} \toprule
\textbf{Papers}            & \multicolumn{1}{l}{\textbf{\begin{tabular}[c]{@{}l@{}}Alternative \\ specific \\ attributes\end{tabular}}} & \multicolumn{1}{l}{\textbf{\begin{tabular}[c]{@{}l@{}}Monotonicity \\ constraint\end{tabular}}} & \multicolumn{1}{l}{\textbf{\begin{tabular}[c]{@{}l@{}}Intrinsically \\ interpretable \\ utility \\ function\end{tabular}}} & \multicolumn{1}{l}{\textbf{\begin{tabular}[c]{@{}l@{}}Automatic \\ idenfication of \\ non-linearities\end{tabular}}} &\multicolumn{1}{l}{\textbf{\begin{tabular}[c]{@{}l@{}}Non-i.i.d \\ error \\ terms\end{tabular}}} \\ \midrule
\citet{wang_deep_2021}         &      X                                                                                                                                                                                                                                                 &                                                                                                                                                                                      &    &     &                                                                                                \\ \cmidrule(lr){2-6}
\citet{martin2021revisiting}      &    X                                                                                                                                                                                                                                                 &                                                                                                                                                                        &   &           &                                                                                          \\ \cmidrule(lr){2-6}
\citet{wong_reslogit_2021}     & X                                                                                                                                                                                                                                                 &                                                                                                                                                                            & $\sim$   &          &   X                                                                                        \\ \cmidrule(lr){2-6}
\citet{sifringer_enhancing_2020}  & X                                                                                                                                                                                                                                                    &                                                                                                                                                                              &  $\sim$  &         &   X                                                                                         \\ \cmidrule(lr){2-6}
\citet{han2022neural}  & X                                                                                                                                                                                                                                               & X                                                                                                                                                                               & X   &                &                                                                                     \\ \cmidrule(lr){2-6}
\citet{kim2023new}  & X                                                                                                                                                                                                                                                             & X                                                                                                                                                                               &  $\sim$  & $\sim$          &                                                                                          \\ \cmidrule(lr){2-6}
\citet{krueger_stated_2022} & X                                                                                                                                                                                                                                                           & X                                                                                                                                                                              &  $\sim$   &           & X                                                                                          \\ \cmidrule(lr){2-6}
\citet{aboutaleb2022theory}  & X                                                                                                                                                                                                                                                        & X                                                                                                                                                                                 & X   &      & X                                                                                               \\ \cmidrule(lr){2-6}
\citet{ortelli_assisted_2021}      & X                                                                                                                                                                                                                                                         &                                                                                                                                                            X          &  X  &                      &   X                                                        \\ \bottomrule
\end{tabular}

\label{contribution_summary}
\end{table}

Whilst the above review evaluates the literature in hybrid data-driven utility modelling, it is also worth considering the more general field of eXplainable Artificial Intelligence (XAI), where researchers aim to add explanations to the predictions of general purpose ML algorithms. The most well known XAI techniques, in use across multiple fields, are:
LIME \citep{ribeiro2016should} and SHAP \citep{lundberg2017consistent}. Several 
researchers use them to provide interpretability of ML classifiers in a mode choice 
context \citep{TAMIMKASHIFI2022279, ren2023exploring, dahmen2023interpretable, martin-baos_prediction_2023}. 
However these methods are local approximations of the model relying on data points 
and, therefore, have no guarantee of validity outside of the input space. 
On the other hand, Explainable Boosting Machine (EBM) from \citet{caruana2015intelligible} is an example of interpretable 
Gradient Boosting Decision Trees algorithm. They achieve interpretability by growing shallow regression trees. However, the model does not allow the modeller to specify alternative-specific variables and monotonicity is only applied as a post-processing tool on features without interaction. In addition, a feature can interact with several other features simultaneously, which is inconsistent with random utility theory. We also observe that the majority of research on ML models for choice modelling focuses
on neural networks. Although neural networks have numerous applications in many fields and
show great results, other algorithms exhibit better performance on classification
tasks such as choice modelling. In particular, GBDT models, including 
XGBoost \citep{chen_xgboost_2016} and LightGBM \citep{ke2017lightgbm} have 
demonstrated state-of-the-art predictive performance for a variety of classification tasks 
\citep{hillel_recreating_2018, martin-baos_prediction_2023}. These models, despite 
lacking of behaviour interpretability, consistency and robustness (like other ML 
models) show great interpretability potential due to their additive nature.
Furthermore, existing ML approaches have continued to rely on the logit formulation, with i.i.d. EV error terms. As such, they cannot capture complex behavioural phenomenons (e.g. 
panel effects, dynamic or sequential choice situations, nesting of alternatives, etc).

In this paper, we present \emph{Random Utility Models with Boosting} (RUMBoost) which aims to combine the predictive power of GBDTs with the interpretability and 
behavioural consistency of DCMs.
At a high level, RUMBoost replaces each parameter in the utility specifications of 
traditional DCMs with an ensemble of regression trees, allowing for non-linear parameters
to be extracted directly from data. 
Algorithmically, RUMBoost consists of two parts:
\begin{enumerate*}[label=(\roman*)]
    \item \emph{Gradient Boosted Utility Values} (GBUV), where ensembles of regression trees are used to impute piece-wise constant values for each parameter in each pseudo-utility specification; and 
    \item \emph{Piece-wise Cubic Utility Functions} (PCUF), where monotonic piece-wise cubic splines are optimised to fit the GBUV outputs, to allow for a defined gradient for each parameter where elasticities are needed. 
\end{enumerate*}

The GBUV output effectively defines piece-wise constant pseudo-utility specifications that are consistent with a DCM, with the ability to constrain the model to have:
\begin{enumerate*}[label=(\roman*)]
    \item pseudo-utility functions depending only on their alternative-specific attributes and socio-economic characteristics;
    \item monotonicity of marginal pseudo-utilities; and
    \item arbitrary interactions between variables as defined by the modeller. 
\end{enumerate*}
Furthermore, the GBUV output is intrinsically interpretable, with the exact functional form observable to the modeller over the full input space of each variable. Hereafter, we no longer distinguish between utility and pseudo-utility as we believe that, with the constraints aforementionned, the GBUV output can be interpreted as utility values.

As the GBUV output is piece-wise constant, it does not have a defined gradient, with the gradient being either zero or infinite at any given point. As such, behavioural indicators such as elasticities and VoT cannot be extracted directly. 
This is addressed through the second element of RUMBoost, PCUF. PCUF only needs to be applied for variables for which elasticities are needed (typically the alternative-specific attributes), which optimises 
a set of piece-wise cubic splines (polynomials) to obtain a defined gradient over the full range of the input 
space. This allows the modeller to compute behavioural indicators such as the Value 
of Time (VoT). 

The flexibility of our approach allows for estimating 
complex model structures with correlated error terms such as the Nested Logit model 
or accounting for population heterogeneity such as the Mixed Logit model. The model is implemented with Python and the code is 
freely available on GitHub (\url{https://github.com/NicoSlvd/RUMBoost}).

The rest of the paper is organised as follows: Section \ref{Methodology} presents the methodology of this 
paper, the results of a case study are presented in 
Section \ref{CS: LPMC}, Section \ref{model extensions} provides some model extensions, and Section \ref{Conclusion and further 
work} concludes and suggests areas of improvement for future work.

\section{Methodology}
\label{Methodology}

\subsection{Theoretical background}

We summarise here key theory of both RUMs and GBDTs under a unified notation in order to help the reader understand our approach. For a more detailed background on each approach, we direct the reader to the following texts:
\citet{Ben-Akiva_Lerman_1985, train2009discrete, friedman_greedy_2001, chen_xgboost_2016, ke2017lightgbm}.

\subsubsection{RUMs}

Discrete Choice Models based on Random Utility theory assume that 
individuals choose the alternative that maximises their utility, a latent
measure of the attractivity of an alternative. The utility is usually 
defined as follows:

\begin{equation}
    U_{in} = V_{in} + \varepsilon_{in}
\end{equation}
where $V_{in}$ is the observable utility and $\varepsilon_{in}$ the error term
capturing unobserved variables for alternative $i$ and individual $n$. In a
DCM, $V_{in}$ is a manually specified linear-in-parameters function of the
variables, such that:

\begin{equation}
    V_{in} = \beta_{i,0} + \sum_{k}^{K} \beta_{i,k} x_{ikn}
\end{equation}
where $K+1$ parameters associated with $K$ variables $x_{ikn}$ have to be estimated 
from the data. For simplicity, we consider here the MNL model, where the error term is 
assumed to be i.i.d, following an extreme value distribution of type 1
(Gumbel). We will later demonstrate more complex model specifications where the
error terms are allowed to be correlated across alternatives. For
the MNL, in a classification problem with $J$ classes,
the probabilities are given by the multi-class logistic function:

\begin{equation}
    P_{in} = \frac{e^{V_{in}}}{\sum_{j}^{J}e^{V_{jn}}}
\end{equation}

The qualities of the MNL model are directly derived from the parametric 
form of $V_{in}$. In particular, it is \textit{intrinsically} interpretable because there
are no variable interactions.

In addition, the modeller can easily incorporate domain knowledge through 
alternative-specific attributes and monotonicity of marginal utilities.
For a multi-class classification problem with $J$ classes, and $K$ 
explanatory variables, we define: 

\begin{equation}
    \bm{x}_i = \bm{a}_i \cup s
\end{equation}
where $x_i$ is the set of variables for alternative $i$ and

\begin{enumerate}[label=(\roman*)]
    \item  $|\bm{x}_i| \leq K$, i.e. the cardinality of $\bm{x}_i$ should be smaller or equal than the number of attributes $K$
    \item $\bm{a}_i$ is the set of alternative-specific attributes for alternative $i$, such that $\bm{a}_i \in \bm{x}, \ \forall i$ and $\bm{a}_i \cap \bm{a}_j = \emptyset, \ \forall i \neq j$
    \item  $\bm{s}$ is the set of socio-economic characteristics such that $\bm{s} \in \bm{x}$ and $\bm{s} \cap \bm{a}_i = \emptyset, \ \forall i$
\end{enumerate}

Discrete choice models also guarantee the monotonicity of marginal utilities.
A positive (resp. negative) monotonic relationship implies that an increasing
attribute $\bm{x}_k$ will increase (resp. decrease) the value of the utility function 
$V_{in}$. The signs of parameters in DCMs allow the modeller to easily 
verify the monotonicity of an attribute and, if needed, to constrain it 
with parameter bounds.

Finally, the gradient is always defined and easy to compute, which allows
for the derivation of behavioural indicators. Among them, the Value of 
Time (VoT) is an important measure of the perceived cost of travelling 
time by individuals. The VoT of an alternative $i$ is defined as follows:

\begin{equation}
    VoT_i = \frac{\partial V_i}{\partial x_{i,time}} \cdot \frac{1}{\frac{\partial V_{i}}{\partial x_{i,cost}}} = \frac{\beta_{i,time}}{\beta_{i,cost}}
\end{equation}
where $\beta_{i,time}$ and $\beta_{i,cost}$ are the parameters associated 
with the time $x_{i,time}$ and cost $x_{i,cost}$ attributes of the utility
function $V_i$.

\subsubsection{Gradient boosting}

The following notation combines that of \citet{friedman_greedy_2001} and \citet{chen_xgboost_2016}, summarising the GBDT algorithm commonly used in popular libraries such as XGBoost and LightGBM. Deep learning models, including DNNs and GBDT use the same multi-class 
logistic function (typically referred to as the softmax function) as the 
MNL, which means that they are based on the same assumptions on 
the error term. In other words, these models first estimate a latent 
regression value for each class which is then used to generate choice 
probabilities. The advantage of these models over a linear-in-parameter
RUM is that they are inherently non-linear, and so can capture complex 
relationships between the input features and the choice probabilities. 
However, unlike in RUMs, these regression models:
\begin{enumerate*}[label=(\roman*)]
    \item are a function of all explanatory variables/features (i.e. attributes of each alternative);
    \item do not contain any constraints to represent behavioural assumptions; 
    \item allow for complex feature interactions that cannot be constrained (or in many cases even observed) by the modeller; and
    \item have an unknown functional form that is not observable by the modeller. 
\end{enumerate*}
Formally, the ML predictive function $F_i(\textbf{x})$ is used to replace 
the deterministic part of the utility function $V_i(\textbf{x})$. In GBDT 
algorithms, the predictive function is an additive function of the form:

\begin{equation}
    V_i(\textbf{x}) = F_{i, m}(\textbf{x}) = \sum_{m} f_{i,m}(\textbf{x})
\end{equation}
where $f_{i,m}(\textbf{x})$ is the output of a single regression tree. At each boosting 
round $m$, assuming $J$ classes, $J$ regression trees are induced to directly minimise
the following objective:

\begin{equation}
    L = \sum_{n=1}^{N}\sum_{i=1}^{J}\ell(y_{in}, \hat{y}_{in, m-1} +  f_{i,m}(\bm{x}_{in}))
\end{equation}
where:
\begin{itemize}
    \item $N$ is the number of observations in the dataset;
    \item $y_{in} = 1$ if the choice $i_n$ of the individual $n$ is $i$, and $0$ otherwise; and
    \item $\hat{y}_{in, m-1}$ is the predicted probability of class $i$ and observation $n$ with attributes $\bm{x}_{in}$ at iteration $m-1$, \textit{i.e.} $\hat{y}_{in, m-1} = F_{i,m-1}(x_{in})$.
\end{itemize}

This optimisation problem has no closed-form solution. It is generally solved by 
taking the Taylor second-order approximation of the loss function:

\begin{equation}\label{newton_raphston}
    L = \sum_{n=1}^{N}\sum_{i=1}^{J} \ell(y_{in}, \hat{y}_{in, m-1}) + g_{in} f_{i,m}(\bm{x}_{in}) + \frac{1}{2} h_{in} f_{i,m}(\bm{x}_{in})^2
\end{equation}
where $g_{in} = \partial\ell(y_{in}, \hat{y}_{in, m-1})/\partial\hat{y}_{in, m-
1}$ and $h_{in} = \partial^2\ell(y_{in}, \hat{y}_{in, m-1})/
\partial^2\hat{y}_{in, m-1}$ are the first and second derivative of the loss function 
with respect to the prediction for observation $n$ and class $i$. Here, the Hessian
matrix is replaced by a diagonal approximation following \citet{friedman_greedy_2001}, 
which means that each class prediction is assumed to be independent. This approximation is crucial for 
computational efficiency and allows for efficient optimisation of the loss function. In 
addition, we can ignore the constant term $\ell(y_{in}, \hat{y}_{in, m-1})$ for the minimisation task.

Assuming that we have $L$ terminal nodes (i.e. the bottom nodes of the regression tree) resulting in $L$ regions 
$L_{im}$ for a tree of class $i$ at iteration $m$, each observation will 
belong uniquely to one of the region $L_{im}$, such that Equation 
\ref{newton_raphston} becomes:

\begin{equation}\label{loss with grad}
    L = \sum_{i=1}^{J} \sum_{l \in L_{im}} (\sum_{n \in l}g_{in}) \gamma_{l,im} + \frac{1}{2} (\sum_{n \in l}h_{in}) \gamma_{l,im}^2
\end{equation}
with $\gamma_{l,im}$ being the leaf value at region $l$ for class $i$. Therefore, by taking the first derivative with respect to $\gamma$ and setting it to $0$, we obtain the optimal leaf value:

\begin{equation}\label{leaf}
    \gamma_{l,im} = - \frac{(\sum_{n \in l}g_{in})}{(\sum_{n \in l}h_{in})}
\end{equation}

By substituting Equation \ref{leaf} into \ref{loss with grad}, and assuming 
a split point that would lead to a left $L_{left,im}$ and right 
$L_{right,im}$ regions, we obtain the loss reduction for splitting:

\begin{equation}\label{loss split}
    L_{split} = \frac{1}{2} \sum_{i=1}^{J} \left( \sum_{l_{left} \in L_{left,im}} \frac{(\sum_{n \in l_{left}} g_{in})^2}{\sum_{n \in l_{left}}h_{in}} + \sum_{l_{right} \in L_{right,im}} \frac{(\sum_{n \in l_{right}} g_{in})^2}{\sum_{n \in l_{right}}h_{in}} - \sum_{l \in L_{im}} \frac{(\sum_{n \in l} g_{in})^2}{\sum_{n \in l}h_{in}}\right)
\end{equation}

Therefore, we can choose the split point that maximises the loss reduction.
For a classification problem, the loss function is usually defined as:

\begin{equation}\label{loss_function}
    \ell = y_{in}\cdot \log(p_{i}(\bm{x_{in}})),
\end{equation}
where $p_{i}(\bm{x}_{in}) = \hat{y}_{in}$. This equation is typically referred to as the cross-entropy loss in ML contexts, though is equivalent to the log likelihood in RUMs.  As in the MNL, we can obtain the predicted probabilities of each class using the softmax function:

\begin{equation}\label{softmax}
    p_{i}(\bm{x}_{in}) = \frac{e^{V_{i}(\bm{x}_{in})}}{\sum_{j=1}^{J}e^{V_{j}(\bm{x}_{jn})}},
\end{equation}

Plugging Equations \ref{loss_function} and \ref{softmax} into Equation 
\ref{leaf} gives:

\begin{equation}
    \gamma_{l,im} = \frac{J-1}{J}\frac{\sum_{\bm{x}_{in}\in R_{l,im}}y_{in}-p_{i}(\bm{x}_{in})}{\sum_{\bm{x}_{in}\in R_{l,im}}p_{i}(\bm{x}_{in})(y_{in}-p_{i}(\bm{x}_{in}))}
\end{equation}
where $\frac{J-1}{J}$ is a factor accounting for redundancy.

The utilities of each ensemble for each individual can then be updated at each iteration:
\begin{equation}
    V_{i,m}(x_{in}) = V_{i,m-1}(\bm{x}_{in}) + \sum_{l=1}^{L}\gamma_{l,im} \cdot \mathbf{1}(\bm{x}_{in}\in L_{im})
\end{equation}
where $\mathbf{1}(\cdot)$ is the indicator function, i.e. equals $1$ if the argument is true, $0$ otherwise.

Monotonicity is easily implemented in regression trees. If we assume a positive (resp. negative) monotonic relationship for an explanatory variable, a split point that partitions the data in two such that $\mathbf{x}_{left} < \mathbf{x}_{right}$ is only considered if:
\begin{equation}
    \gamma_{left} < \gamma_{right} \ \ \ \ \text{(resp.} \ \gamma_{left} > \gamma_{right}\text{)}
\end{equation}
Note that, in order to satisfy the monotonicity over the full tree, the subsequent left and right leaf values are bounded by $\frac{(\gamma_{left}+\gamma_{right})}{2}$. The side of the tree and the nature (positive or negative) of the constraint determine if it is a lower or upper bound.

\subsection{The RUMBoost model}

We first explain here how we adapt the general GBDT model to output Gradient Boosted Utility Values (GBUV) to emulate parametric RUMs. We then present the Piecewise-Cubic Utility Functions (PCUF) algorithm, that outputs smoothed monotonic non-linear parameters.

\subsubsection{Gradient Boosted Utility Values (GBUV)}\label{GBUV}

In RUMBoost-GBUV, we replace each parameter in the utility functions of a RUM with an ensemble of regression trees, where the leaves in the regression trees represent the partial utility contribution for the corresponding value of that variable. These can then be added over each tree in the ensemble to find the contribution of each variable to the utility. The overall utility for each alternative can therefore be found by summing the ensembles for each variable over all variables in the utility function. For K parameters applied to K variables, we have:

\begin{equation}
    V_{in} = \text{ASC}_{i} + \sum_{k}^{K_i} \sum_{m}^{M_{ik}} f_{imk} (x_{ink})
\end{equation}
where $\text{ASC}_i$ is an Alternative-Specific Constant for alternative $i$ and $M_{ik}$ is the number of regression trees in the ensemble for parameter $k$ for alternative $i$. In other words, by ordering the split points for each variable in ascending order, and adding leaf values for the appropriate values, we can interpret the sum as raw utility values for each variable. This allows us to impute the full functional form of the "non-linear parameters" with piece-wise constants. In this form, the utility function is non-linear (and non-continuous) and known over the full input space. Probabilities for each alternative can then be calculated with the appropriate transformation (e.g. softmax\slash logistic function for the MNL model). 

For simplicity of notation, the above formulation shows each parameter being applied to a single variable. However, it is possible to define arbitrary feature interactions by allowing the regression trees in each ensemble to split on multiple variables e.g. for an interaction of two variables $k_1$ and $k_2$ we would have $\sum_{m}^{M_{ik}} f_{imk}(x_{in{k_1}}, x_{in{k_2}})$. This allows the modeller to specify any desired feature interactions as required. 

We modify the standard GBDT boosting algorithm in order to identify optimal RUMBoost-GBUV ensembles for each variable (or combination of variables). For a J-class problem, we introduce J trees at each boosting iteration during model training, one for the utility function of each alternative. At each iteration, the model computes choice probabilities for the previous estimates of the utility functions, derives the gradient and hessian of the log-loss, and uses them for boosting the next set of $J$ trees. Each regression tree is added to the ensemble for a single variable, automatically selected by the algorithm in an exhaustive search (for computational efficiency the search is typically limited to a finite number of possible split points) in order to minimise the loss function.\footnote{Note that for the practical implementation of RUMBoost-GBUV, there is actually a single ensemble per alternative utility function, with each tree in the ensemble restricted to only split on the corresponding variable(s). The trees for each variable are then grouped into a separate ensemble once the model has been fully trained. However, this is equivalent to having a separate ensemble for each variable during model training, which we believe to be a more intuitive interpretation of the underlying functionality, and so present the algorithm in that way here.}

Once the model has been fully trained, we can extract normalised ASCs. We define the non normalised ASCs as:

\begin{equation}
    \text{ASC}'_i = \sum_{k}^{K_i} \sum_{m}^{M_{ik}} f_{imk} (0)
\end{equation}

Since in RUM only the difference in utilities matter, one of the ASCs can be normalised to $0$. Assuming that the ASC of the alternative~$j$ is normalised, we
obtain the following set of ASCs: 

\begin{equation}\label{ASC}
    \text{ASC}_i = \text{ASC}'_i - \text{ASC}'_j, \forall i=1,...,J
\end{equation}

We build our algorithm on top of LightGBM \citep{ke2017lightgbm}, such that each ensemble is a LightGBM Booster object. Therefore, we can make use of the already implemented monotonicity constraint feature to impose monotonicity on utilities as required. 

The RUMBoost-GBUV model training algorithm is described formally in Algorithm \ref{alg:rumboost}. Note that the algorithm is independent of the assumption on the error terms. This allows RUMBoost-GBUV to be used to emulate any arbitrary model formulation for which the gradient and Hessian of the loss function is defined, including Nested Logit and Mixed Logit models. Furthermore, as the code has been re-implemented from scratch, any ML regressor that can satisfy the constraints described above can be used in place of GBDT.

\begin{algorithm}[H]
\caption{\bfseries: RUMBoost-GBUV}\label{alg:rumboost}
\begin{algorithmic}
\State $\bm{x_{i}} = \bm{a_i} + \bm{s}$,\ \ \ \ $\forall i = 1, ..., J$
\State Positive monotonic set of attributes $\bm{x}_i^+ \subseteq \bm{x}_{i}$, \ \ \ \ $\forall i = 1, ..., J$
\State Negative monotonic set of attributes $\bm{x}_i^- \subseteq \bm{x}_{i}$, \ \ \ \ $\forall i = 1, ..., J$
\State $\bm{x}_i^+ \cap \bm{x}_i^- = \emptyset$, \ \ \ \ $\forall i = 1, ..., J$
\State Specify variables interactions
\State $V_{in}(\bm{x_{in}}) = 0$,\ \ \ \ $\forall i = 1, ..., J$
\For{$m = 1$ to $M$}
    \State  Compute $p_{in}(\bm{x}_{in})$
    \For{$i = 1$ to $J$}
        \State Compute the first and second derivatives of the loss function
        \State Choose the split point that maximises the loss reduction of any variable $k$ 
        \State Add the regression tree $f_{imk}(x_{ink})$ to its corresponding ensemble
    \EndFor
\EndFor
\end{algorithmic}
\end{algorithm}

\subsubsection{Piece-wise Cubic Utility Function (PCUF)}\label{splines}

The GBUV ensembles for each parameter in Section \ref{GBUV} are non-continuous, and so have a gradient of either zero or infinite at any point. However, many behavioural indicators require the utility 
function to have defined gradient to be computed. Therefore, we interpolate the utility 
values into a smooth function using piece-wise cubic Hermite
splines. Using splines ensure that the underlying cubic polynomials 
have equal values and derivatives at the \emph{knots} (i.e. boundary points). Given the number of 
knots and their positions, only the derivative at each knot needs to be computed, 
making Hermite splines very attractive for efficient computations. Using the approach introduced by \citet{fritsch1980monotone}, it is possible to guarantee monotonic splines, where the gradient is always negative or positive (or zero) as required. 
The interpolation must satisfy two conflicting objectives:
\begin{enumerate*}[label=(\roman*)]
    \item fitting the data as well as possible to maintain good predictive power on out-of-sample data; and
    \item being as smooth as possible to obtain relevant behavioural indicators.
\end{enumerate*}

The first objective favours a higher number of knots, while the second aims for a lower
number so that the derivative is well defined. A natural objective 
function to capture the trade-off of both these objectives is the Bayesian 
Information Criterion (BIC), which takes the following form:

\begin{equation}
BIC = -2N \cdot L + df\cdot \ln(N)
\end{equation}
where $L$ is the loss function described in Equation \ref{loss_function}, $df$ is the degree 
of freedom of the model, and $N$ is the number of observations. The first part of the function aims
for a better fit of the data, while the second part penalises the model for its complexity. 

RUMBoost-PCUF, therefore, has two parameters to tune: 
\begin{enumerate*}[label=(\roman*)]
    \item the number of knots; and
    \item their positions.
\end{enumerate*}
Given a sequence of $Q + 1$ knots 
$a_k = t_{0,k} < t_{1,k} < ... < t_{Q, k} = b_k$ for an attribute $k$ where $a_k$ and $b_k$ are the domain
where that attribute is defined, the optimal positions and numbers of knots are determined by 
the following optimisation problem:

\begin{equation}
\label{optimisation}
\begin{aligned}
\min_{t_{q,k}} \quad & -2N \cdot L + df\cdot ln(N)\\
\textrm{s.t.} \quad & t_{q+1, k}-t_{q, k} > 0 & \forall q=0, ..., Q, \forall k \\
  &  t_{0, k} = a_k & \forall k \\
  &  t_{Q, k} = b_k & \forall k \\
\end{aligned}
\end{equation}

Given the number of knots, there is an optimal position of knots that minimises the loss
function. Therefore, the two hyperparameters can be tuned sequentially: the number of knots is selected first
and their optimal positions are found with a constrained optimisation solver afterwards. 
However, this can be a complex optimisation problem if the number of attributes is high, and 
it has been shown in the literature that it is acceptable to fix the positions of knots 
over the range of the attribute (e.g., quantile) and optimise only their numbers \citep{hastie1990generalized}.
The RUMBoost-PCUF algorithm is summarised in Algorithm \ref{alg:PCUF}.

\begin{algorithm}[ht]
\caption{\bfseries: RUMBoost-PCUF}\label{alg:PCUF}
\begin{algorithmic}
\State $\bm{x_{i}} = \bm{a_i} + \bm{s}$,\ \ \ \ $\forall i = 1, ..., J$
\State Gradient boosted utility values $V_{ik}$  \ \ \ \ $\forall i = 1, ..., J$, $k = 1,...,K$
\State Number of iterations for hyperparameter search $N_{iter}$
\For{$n=1$ to $N_{iter}$}
    \For{$i = 1$ to $J$}
        \For{$k = 1$ to $K$}
            \If{$x_{k} \in \bm{a_i}$}
                \State Choose a number of knots $Q_k$
                \State Define the initial position of knots $t_{q, k} = Q_k\text{-quantile}(x_{k}, q)$\ \ \ \ $\forall q = 0, ..., Q_k$
            \EndIf
        \EndFor
    \EndFor
    \State Optimise the knots position to minimise the $BIC$
\EndFor
\end{algorithmic}
\end{algorithm}

\subsection{Code and implementation}

We implement the model in Python, making use of the library LightGBM for the utility 
regression ensembles \citep{ke2017lightgbm}. Our implementation creates a regression ensemble 
for each alternative in which the input attributes (or features) can be specified independently. 
We set up the logistic/softmax function so that in each round of boosting, the trees 
(and split points in each tree) are selected to directly minimise the log-loss 
(cross-entropy loss), therefore emulating maximum likelihood estimation (MLE). 
Separate ensembles for each parameter are obtained by restricting the possible set of feature interactions in each tree. The code makes use of the existing monotonic constraints functionality to guarantee monotonicity of the marginal utilities as required. 
We have therefore implemented an interface which allows the modeller to
specify:

\begin{itemize}
    \item which attributes should be included in each utility function
    \item control attribute interactions
    \item specify which attributes should have monotonic marginal utilities. 
\end{itemize}

Early stopping is used to determine the appropriate number of trees in each ensemble, with boosting terminated once the log-loss does not improve on out-of-sample data for a given number of iterations (e.g. 100).
Furthermore, we have written a script that converts model files from the popular choice modelling 
software Biogeme \citep{bierlaire2023short} to be used directly within RUMBoost, therefore allowing
modellers to easily replicate any MNL model in RUMBoost. 
This conversion works using the utility specification to define alternative-specific attributes and 
control attribute interactions, and using bounds on the parameters to define 
monotonic constraint. The code is freely available on Github 
(\url{https://github.com/NicoSlvd/RUMBoost})

\section{Detailed case study}
\label{CS: LPMC}

We apply our methodology on a case study, where we benchmark RUMBoost against a MNL model 
and three ML classifiers: Neural Network (NN), Deep Neural Network (DNN) and LightGBM. These models
are re-implemented from \citet{martin-baos_prediction_2023}\footnote{The code is freely
available at \url{https://github.com/JoseAngelMartinB/prediction-behavioural-analysis-ml-travel-mode-choice}}. In addition, 
we show the non-linear utility functions and compute behavioural indicators. To ensure a fair 
comparison, RUMBoost is built using the same model specification as the MNL model.

\subsection{Case study specifications} \label{datasetLPMC}
We use the London Passenger Mode
Choice (LPMC) \citep{hillel_recreating_2018} dataset for our case study, a publicly available dataset
providing details of more than 80000 trips in London, alongside their 
associated mode choice decisions. It is an augmented version of
the London Travel Demand Survey (LTDS) trip diary dataset, to include the travel
time and cost of alternatives. The dataset contains observations from
17615 households over a three-year period, and there are four possible
alternatives: walking, cycling, public transport and driving. The MNL 
model, also used to create RUMBoost, is a 62-parameter model with alternative-specific constants (ASC). 
When estimating the MNL model, we normalise the ASC, the generic attributes and the socio-economic 
characteristics of the walking alternative to zero. The model specification is
summarised in Table \ref{tab:mnl}. Lastly, the RUMBoost and ML models are trained on the first 
two years of the dataset with a 5-fold cross validation scheme designed in such a way 
that trips performed by the same household members cannot be in different folds, to 
avoid data leakage. We include an early stopping criterion of 100 iterations,
i.e. we stop the training if the performance on the validation set is not improving during 100 iterations.
For the ML classifiers, we also include a hyperparameter search.
This search is done with the python library Hyperopt \citep{bergstra2013making}, 
and the search space and results are summarised in \ref{app:hyperparamers}.

\begin{table}[ht]
\centering
\caption{\centering Variables used in the LPMC RUMBoost and MNL models. For the MNL estimation, the socio-economic characteristics and generic attributes are normalised to 0 for the walking alternative. Purpose and Fuel type are dummy variables where one category is normalised. The constants are not included in the RUMBoost training, but are reconstructed afterwards, following Equation \ref{ASC}.}
\label{tab:mnl}
\begin{tabular}{lcccc} \toprule
\textbf{}                               & \multicolumn{1}{l}{\textbf{Walking}} & \multicolumn{1}{l}{\textbf{Cycling}} & \multicolumn{1}{l}{\textbf{Public Transport}} & \multicolumn{1}{l}{\textbf{Driving}} \\ \midrule
\textit{Alternative-specific attributes} &                                      &                                      &                                               &                                      \\
Constant                                &                                                 & \checkmark            & \checkmark                     & \checkmark            \\
Travel time                             & \checkmark            & \checkmark            & \checkmark                     & \checkmark            \\
Access time                             & \multicolumn{1}{l}{}                 & \multicolumn{1}{l}{}                 & \checkmark                     & \multicolumn{1}{l}{}                 \\
Transfer time                           & \multicolumn{1}{l}{}                 & \multicolumn{1}{l}{}                 & \checkmark                     & \multicolumn{1}{l}{}                 \\
Waiting time                            & \multicolumn{1}{l}{}                 & \multicolumn{1}{l}{}                 & \checkmark                     & \multicolumn{1}{l}{}                 \\
Num. of PT changes                        & \multicolumn{1}{l}{}                 & \multicolumn{1}{l}{}                 & \checkmark                     & \multicolumn{1}{l}{}                 \\
Cost                                    &                                      &                                      & \checkmark                     & \checkmark            \\ 
Congestion charge                                    &                                      &                                      &                     & \checkmark            \\  \\
\textit{Socio-economic characteristics and generic attributes}             &                                      &                                      &                                &                                      \\
Straight-line distance                  &   \checkmark                                    & \checkmark            & \checkmark                     & \checkmark            \\
Starting time                           &   \checkmark                                    & \checkmark            & \checkmark                     & \checkmark            \\
Day of the week                         &   \checkmark                                    & \checkmark            & \checkmark                     & \checkmark            \\
Gender                                  &   \checkmark                                    & \checkmark            & \checkmark                     & \checkmark            \\
Age                                     &   \checkmark                                    & \checkmark            & \checkmark                     & \checkmark            \\
Driving license                         &   \checkmark                                    & \checkmark            & \checkmark                     & \checkmark            \\
Car ownership                           &   \checkmark                                    & \checkmark            & \checkmark                     & \checkmark            \\
Purpose: home-based work                 &   \checkmark                                    & \checkmark            & \checkmark                     & \checkmark            \\
Purpose: home-based education            &   \checkmark                                    & \checkmark            & \checkmark                     & \checkmark            \\
Purpose: home-based other                &   \checkmark                                    & \checkmark            & \checkmark                     & \checkmark            \\
Purpose: employers business              &   \checkmark                                    & \checkmark            & \checkmark                     & \checkmark            \\
Purpose: non-home-based other            &   \checkmark                                    & \checkmark            & \checkmark                     & \checkmark            \\
Fuel type: diesel                         &   \checkmark                                    & \checkmark            & \checkmark                     & \checkmark            \\
Fuel type: hybrid                         &   \checkmark                                    & \checkmark            & \checkmark                     & \checkmark            \\
Fuel type: petrol                         &   \checkmark                                    & \checkmark            & \checkmark                     & \checkmark            \\
Fuel type: average                        &   \checkmark                                    & \checkmark            & \checkmark                     & \checkmark            \\ \bottomrule
\end{tabular}
\end{table}

\subsection{RUMBoost model specification}

The MNL model is directly used to specify the constraints of the RUMBoost model. The
alternative-specific attributes constraint is directly satisfied by the MNL
utility specification. Interactions between attributes are restricted, such that each tree corresponds to a single parameter. Finally, monotonicity constraints
are obtained from the bounds that would be applied to the MNL beta
parameters (see Table \ref{tab:mono}), and so applied negatively on travel time, headway,
cost and distance, and positively on car ownership and driving license (when applicable).

\begin{table}[ht]
\caption{\centering Attributes constrained to monotonicity. Travel time, cost and distance are monotonic negative, i.e. an increase of these attributes will decrease the utility function value. Car ownership and driving license are positive monotonic and applied only on for the driving alternative.}
\centering
\begin{tabular}{cr}\toprule
\textbf{Monotonic negative }                          &  Travel time, cost, distance      \\ \midrule
\textbf{Monotonic positive}  & Car ownership*, driving license*\\
\bottomrule
 & \footnotesize *Only for the driving alternative
\end{tabular} 
\label{tab:mono}
\end{table}

One big advantage of RUMBoost over the more flexible GBDT model is that the additional constraints help to
regularise the model and therefore has a lower propensity to overfit compared to the unconstrained GBDT model. We find that
the modelling results are less dependent on hyperparameter values, including
regularisation parameters. Thus, we use the LightGBM default parameters except for the learning rate, the maximum depth of trees and the number of trees. The number of boosting rounds is obtained with the cross-validation early 
stopping criterion, where we average the best number of trees of each fold
to obtain the final number of trees. As such, we make use of the following parameters for
each regression ensemble:

\begin{itemize}
    \item learning rate: 0.1
    \item max depth: 1 (following attribute interaction constraints)
    \item minimum data and sum of hessian in leaf: 20 (default) and 0.001 (default)
    \item maximum number of bins and minimum of data in bins: 255 (default) and 3 (default)
    \item monotone constraint method: advanced
    \item number of boosting rounds (trees): 1300
\end{itemize}

\subsection{Comparisons with other ML models and MNL}\label{benchmarks}

The results of the RUMBoost model are benchmarked against the MNL and ML models presented
in Section \ref{datasetLPMC}.
We compare the models with their cross-entropy loss (CLE) on the test set 
(lower is better) and their computational time per cross-validation iteration. 
The results are shown in Table \ref{tab:res}.

\begin{table}[ht]
\centering
\caption{\centering Benchmark of Classification on LPMC Dataset. The models are compared with their CEL (negative Cross-Entropy Loss, lower the better) on the test set and their computational time for one CV iteration. The training results of RUMBoost-PCUF are the results of the optimisation problem described in Section \ref{splines}.}
\begin{tabular}{llrr}\toprule
\multirow{2}{*}{\textbf{Models}}        &\multirow{2}{*}{\textbf{Metrics}} & \multicolumn{2}{c}{\textbf{LPMC}}                         \\
                                        &                               & \multicolumn{1}{l}{5 fold CV} & \multicolumn{1}{c}{Holdout test set} \\ \midrule
\multirow{2}{*}{\textbf{MNL}}           & CEL                & 0.6913                         & 0.7085                   \\
                                        & Comp. Time {[}s{]} & 242.14                         & -                        \\ \cmidrule(lr){2-4}
\multirow{2}{*}{\textbf{NN}}            & CEL                & 0.6516                         & 0.6667                   \\
                                        & Comp. Time {[}s{]} & 7.85                           & -                        \\ \cmidrule(lr){2-4}
\multirow{2}{*}{\textbf{DNN}}           & CEL                & 0.6613                         & 0.6735                   \\
                                        & Comp. Time {[}s{]} & 3.89                           & -                        \\ \cmidrule(lr){2-4}
\multirow{2}{*}{\textbf{LightGBM}}      & CEL                & \textbf{0.6381}                & \textbf{0.6537}          \\
                                        & Comp. Time {[}s{]} & 4.64                           & -                        \\ \cmidrule(lr){2-4}
\multirow{2}{*}{\textbf{RUMBoost-GBUV}} & CEL                & 0.6570                         & 0.6737                   \\
                                        & Comp. Time {[}s{]} & 6.48                           & -                        \\  \cmidrule(lr){2-4}
\multirow{2}{*}{\textbf{RUMBoost-PCUF}} & CEL                & 0.6479*                        & 0.6730                   \\
                                        & Comp. Time {[}s{]} & 712.48*                        & -                        \\ \bottomrule
                                        &                    &                                & \footnotesize *Not with CV 
\end{tabular}
\label{tab:res}
\end{table}
Overall, the RUMBoost model outperforms the MNL model on both training and testing validations, whilst still ensuring a directly interpretable functional form. Whilst there is a marginal performance sacrifice of the RUMBoost models compared to the unconstrained GBDT and NN models, it is important to note that the functional form of the RUMBoost utilities is directly interpretable over the full input space, allowing for guarantees of behavioural consistency of forecasts, and in the case of RUMBoost-PCUF, extraction of behavioural indicators, which is not possible for the GBDT and NN models. Note that further enhancements of the RUMBoost model, including the functional effects model (see Section \ref{funct effect rumboost}) further narrow this gap. Interestingly, the loss 
of information due to smoothing is minimal, and the CE loss even 
improves on the LPMC dataset, even with an objective function penalising complex models.
Therefore, we deduce that the piece-wise splines act as further regularisation 
of the RUMBoost model when there is a sufficient amount of data and splitting points.
Computationally, the GBUV algorithm shows similar results than the ML 
classifiers, which is much better than the MNL model. The PCUF algorithm on the other hand 
has a big computational time, illustrating the complexity of the optimisation problem.

\subsection{Gradient boosted utility values}

\begin{figure}[H]
    \centering
    \begin{subfigure}[b]{0.49\textwidth}
        \includegraphics[width=\textwidth]{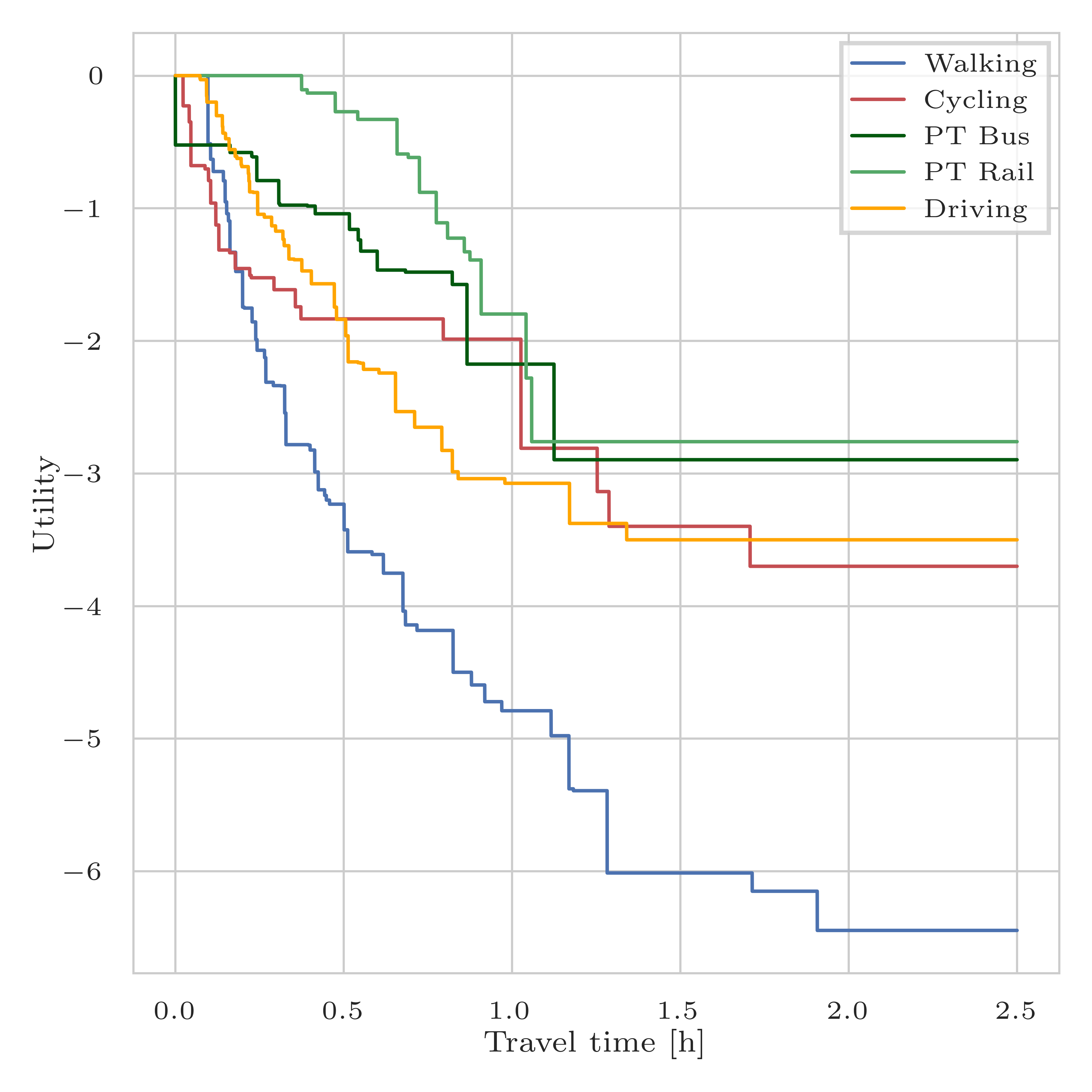}
        \caption{Travel time (LPMC)}
        \label{fig:tt}
    \end{subfigure}
    \begin{subfigure}[b]{0.49\textwidth}
        \includegraphics[width=\textwidth]{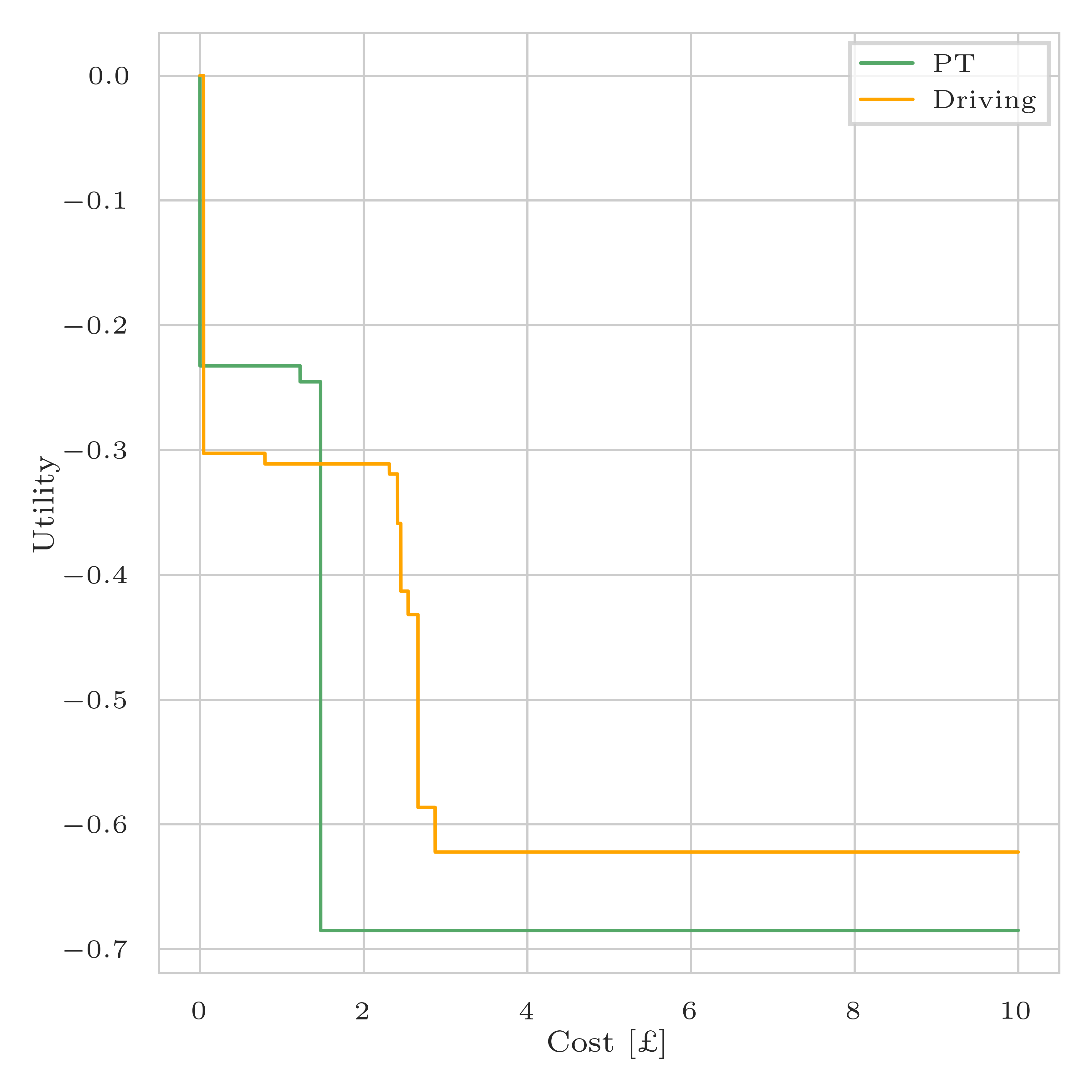}
        \caption{Cost (LPMC)}
        \label{fig:cost}
    \end{subfigure}
    \caption{\centering Utility contributions of a) travel time and b) cost on the LPMC dataset, both under a negative monotonic constraint. Each step represents a split point of a regression tree in the corresponding ensemble.}
    \label{fig:tt+dist}
\end{figure}

\begin{figure}[H]
    \centering
     \begin{subfigure}[b]{0.49\textwidth}
         \centering
         \includegraphics[width=\textwidth]{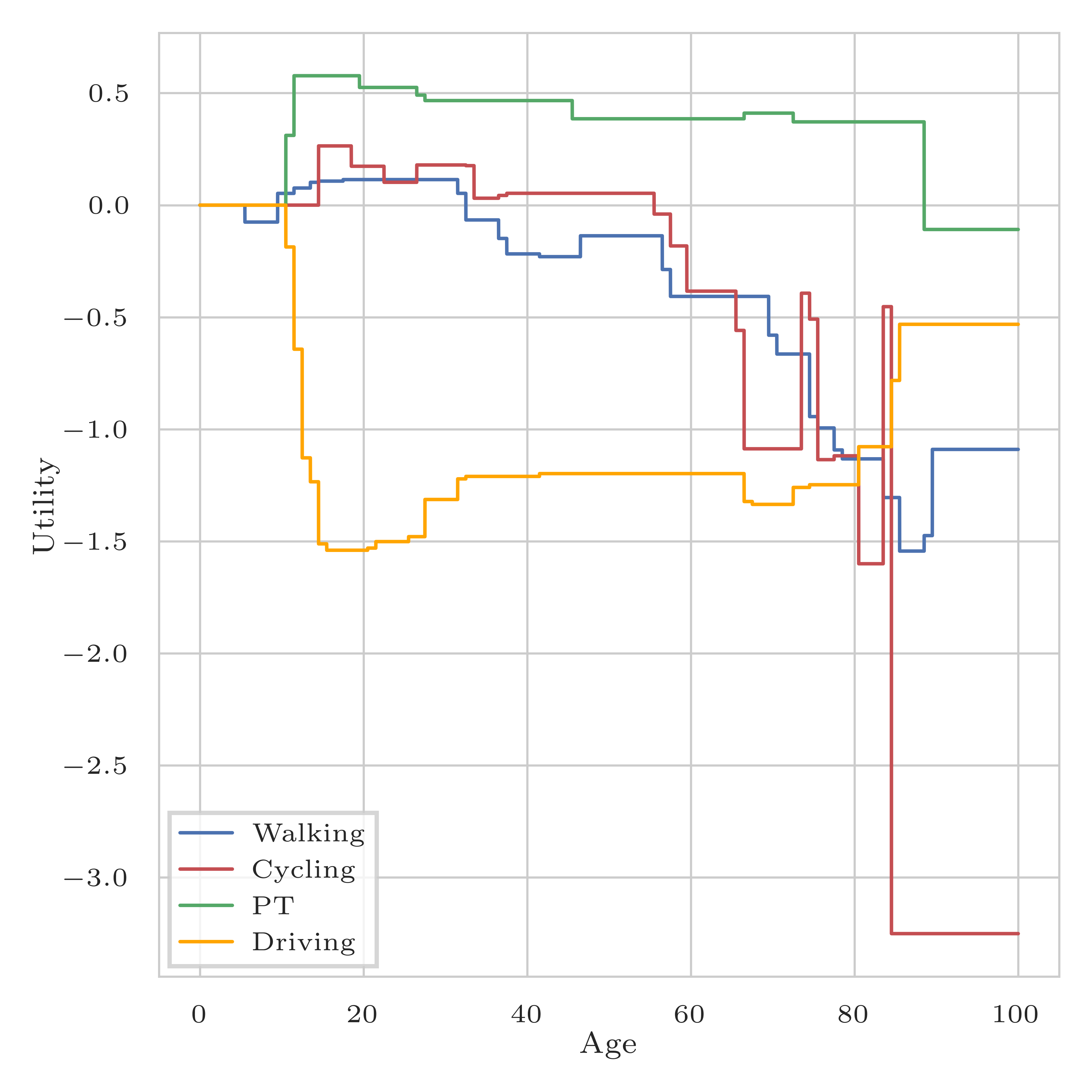}
         \caption{Age (LPMC)}
         \label{fig:age}
     \end{subfigure}
     \begin{subfigure}[b]{0.49\textwidth}
         \centering
         \includegraphics[width=\textwidth]{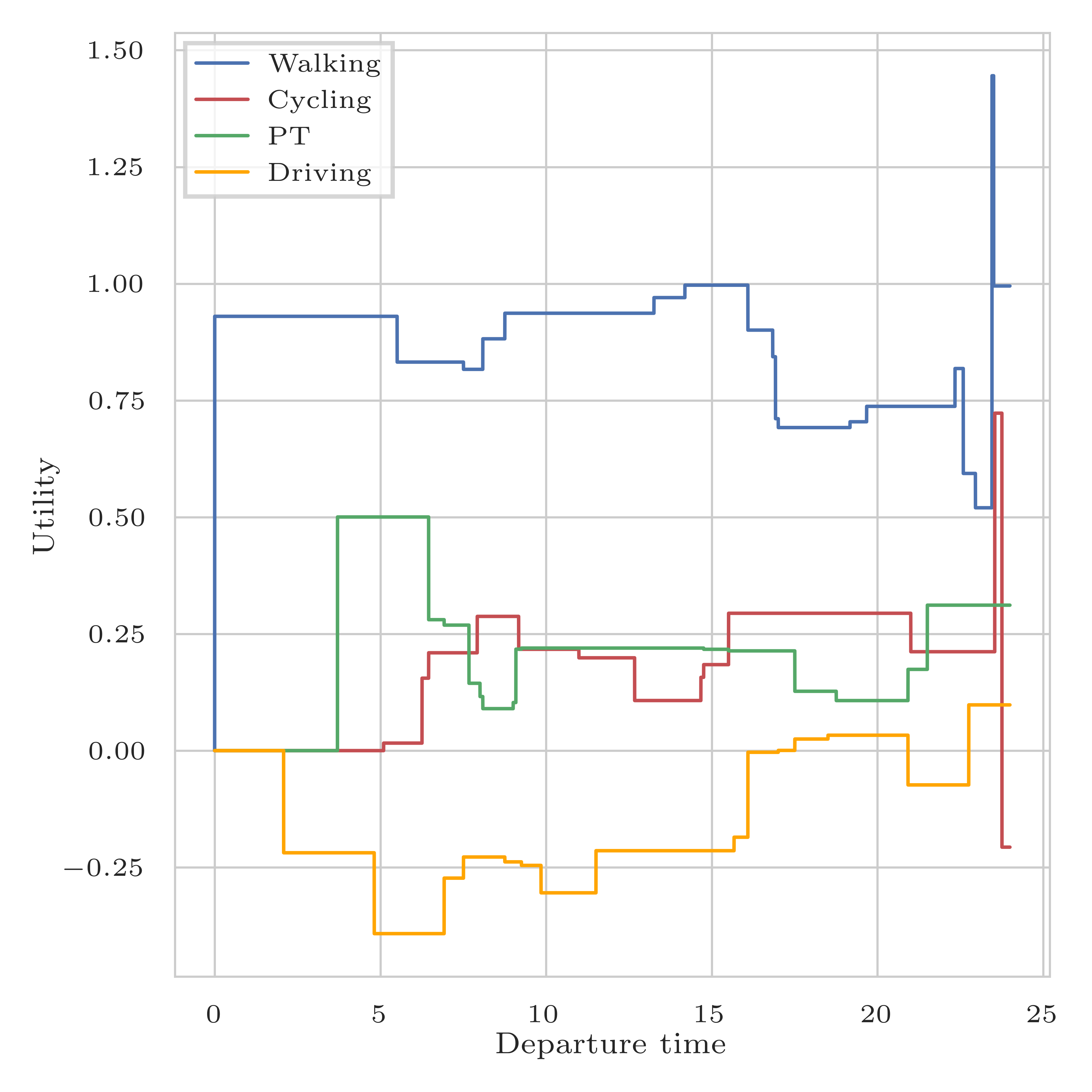}
         \caption{Departure time (LPMC)}
         \label{fig:dep_time}
     \end{subfigure}
    \caption{\centering Utility contributions of a) age and b) departure time on the LPMC dataset. Both variables are non-monotonic. Each step represents a split point of a regression tree in the corresponding ensemble.}
    \label{fig:age+dep_time}
\end{figure}

The primary advantage of using RUMBoost over other unconstrained ML algorithms is that non-linear utility functions can
be observed directly. We have access to the ensemble for each parameter thanks to the custom training
function. Therefore, we can delve into each tree of each ensemble to retrieve
leaf values and splitting points. More specifically, each split point in a regression tree represents a step in the GBUV output for the corresponding parameter. The utility contribution at point $x$ is the cumulative sum of all corresponding leaf values of the trees in the ensemble. These functions are presented for the travel time, cost, departure time and age parameter in Figures \ref{fig:tt}, \ref{fig:cost}, \ref{fig:age} and \ref{fig:dep_time}. 

Figure \ref{fig:tt} shows the impact of travel times on
parameter values for the LPMC dataset, with the public transport alternative
including both bus and train travel times. From the graph, it can be noted that the walking and driving time parameters have a convex shape, indicating that the increase in a short trip has a greater influence on the utility
function than a longer trip. This contrasts with the two parameter values
linked to public transport, where bus travel time has an approximately linear
curve, and train travel time impact results in a concave parameter values. Finally, the parameter values of cycling is initially
convex, then reaches a plateau between 0.5 and 1 hour of travel time,
and decreases rapidly after 1 hour. Figure \ref{fig:cost}
depicts the constant utility contributions of travel cost parameter for driving and PT. Interestingly, 
both parameter values exhibit a similar behaviour.
There is first a sharp decrease representing the disutility of travelling,
then a plateau, and a final drop around 2 £. 

Figures \ref{fig:age} and \ref{fig:dep_time} show the constant utility values of the age and departure time parameters for each alternative. Following behavioural theory, there are no monotonicity constraints on these 
variables, but we still observe that increasing age mainly reduces the utility of walking 
and cycling, with a mostly convex shape. For public transport, the parameter values is the lowest at older ages, 
and peaks around the age of 20. Finally, for the driving alternative, the utility 
is higher with younger and older ages, but the lowest point is at the age of 20. Also note that passengers are included in the driving alternatives,
explaining the high values for younger and older individuals. Finally, the departure time utilities
are stable during the day. The walking and cycling utilities exhibit a sharp increase around
midnight. The public transport
utility has a higher value on the morning, corresponding to the morning commuting time. Finally,
the driving utility decreases until 7h, and increases until the end of the day after that.


\subsubsection{GBUV robustness}

\begin{figure}[ht!]
    \centering
     \begin{subfigure}[b]{0.49\textwidth}
         \centering
         \includegraphics[width=\textwidth]{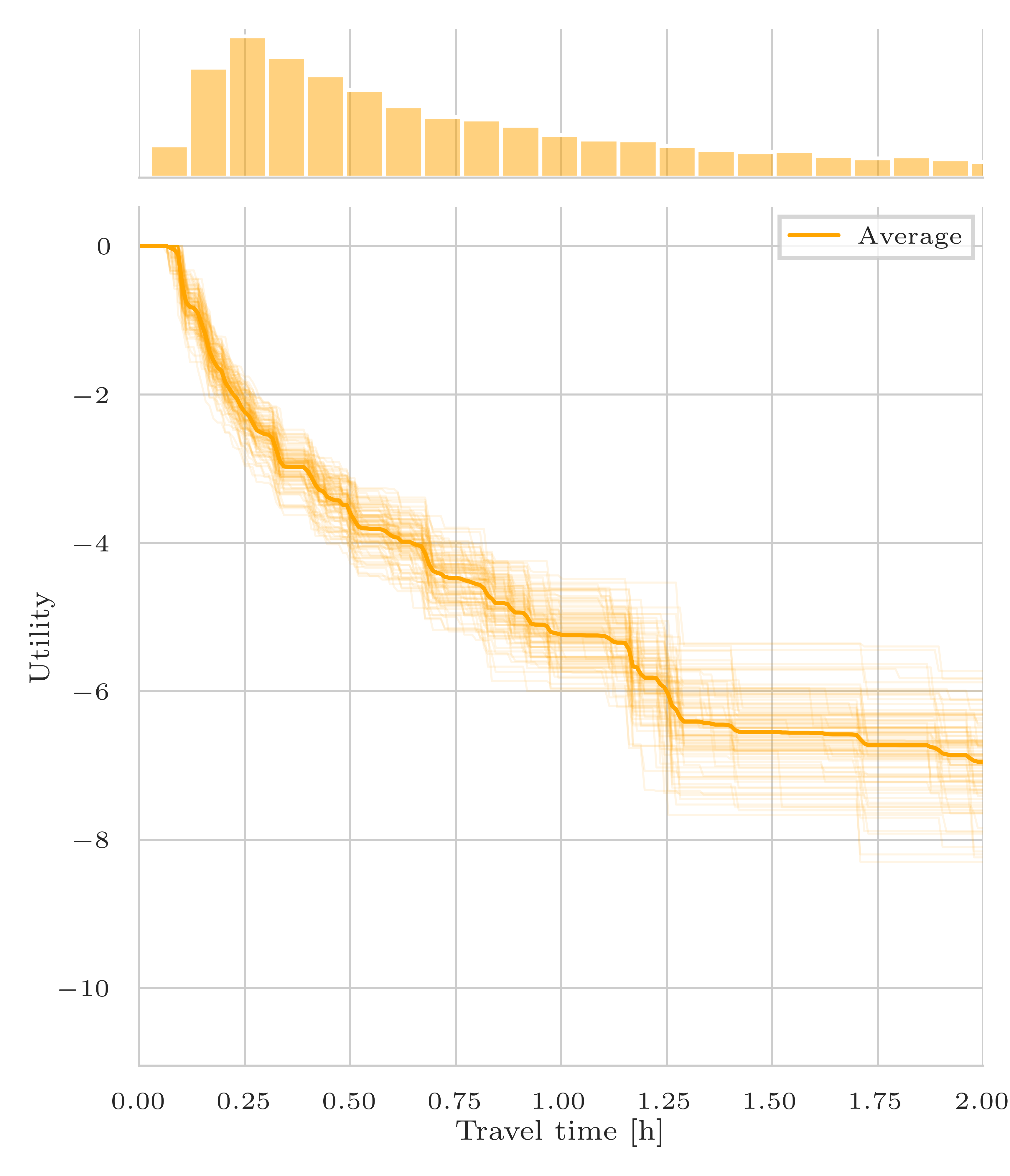}
         \caption{Walking travel time}
         \label{fig:bootstrap_walking}
     \end{subfigure}
     \begin{subfigure}[b]{0.49\textwidth}
         \centering
         \includegraphics[width=\textwidth]{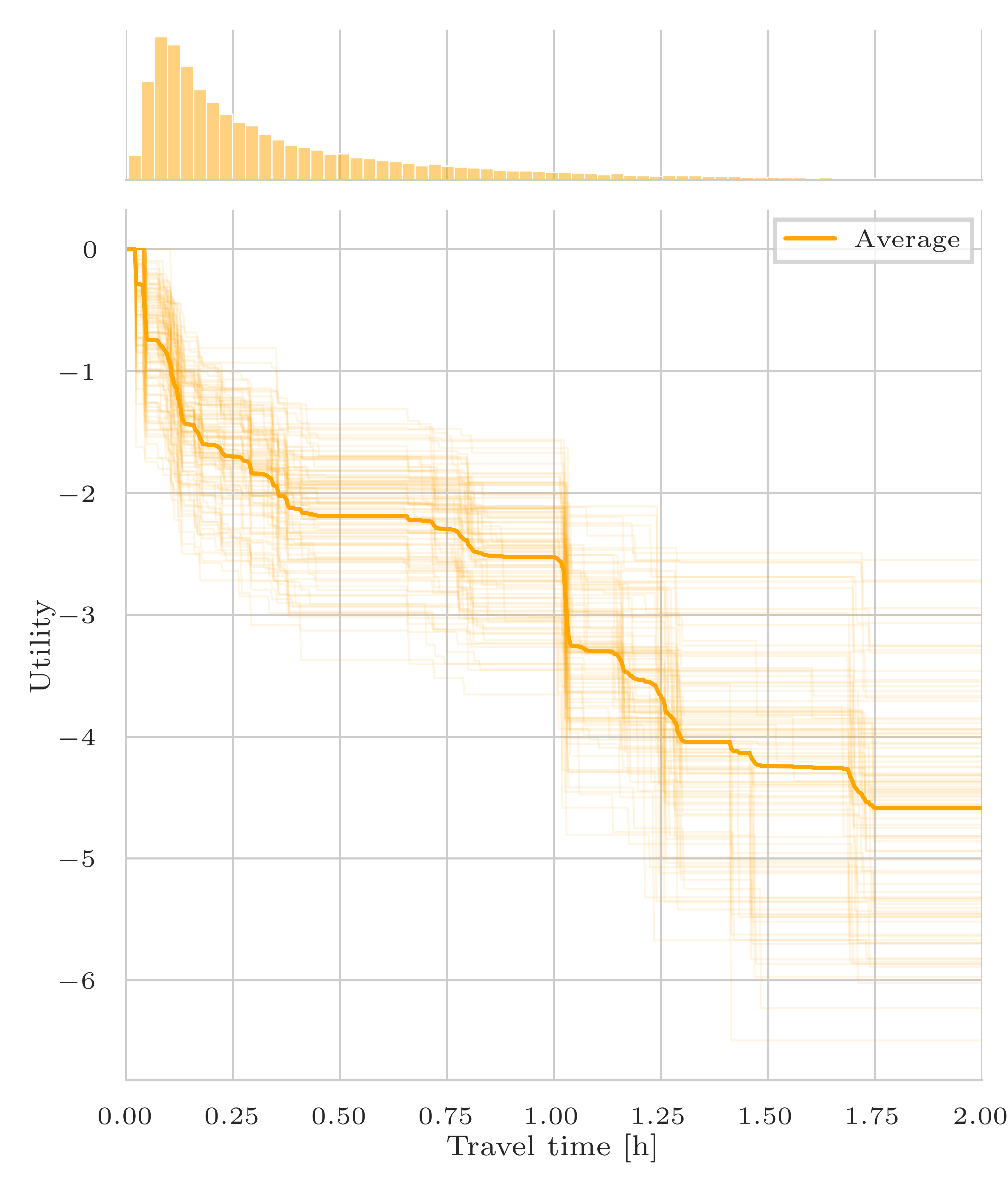}
         \caption{Cycling travel time}
         \label{fig:bootstrap_cycling}
     \end{subfigure}
     \begin{subfigure}[b]{0.49\textwidth}
         \centering
         \includegraphics[width=\textwidth]{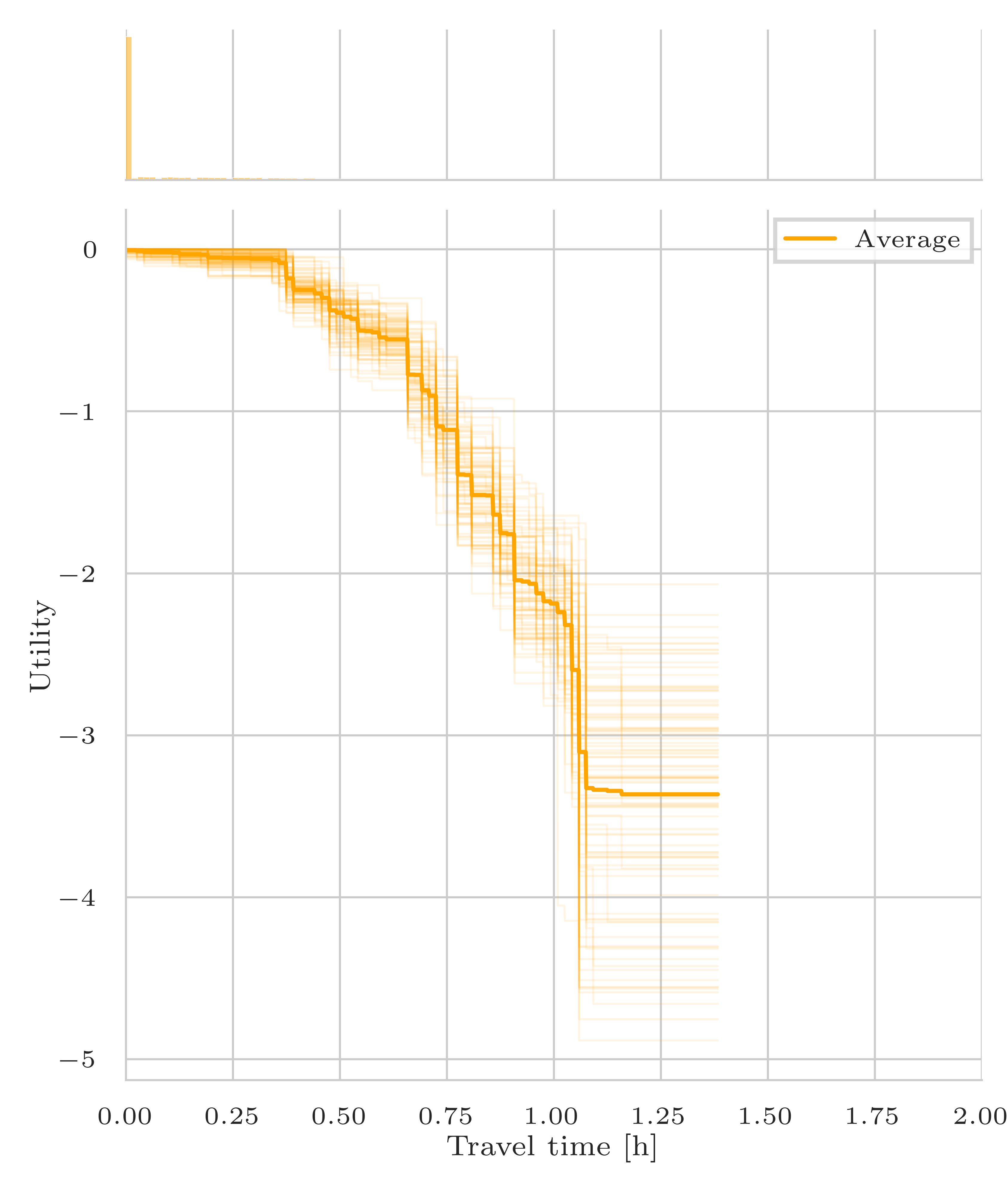}
         \caption{PT - Rail travel time}
         \label{fig:bootstrap_PT}
     \end{subfigure}
     \begin{subfigure}[b]{0.49\textwidth}
         \centering
         \includegraphics[width=\textwidth]{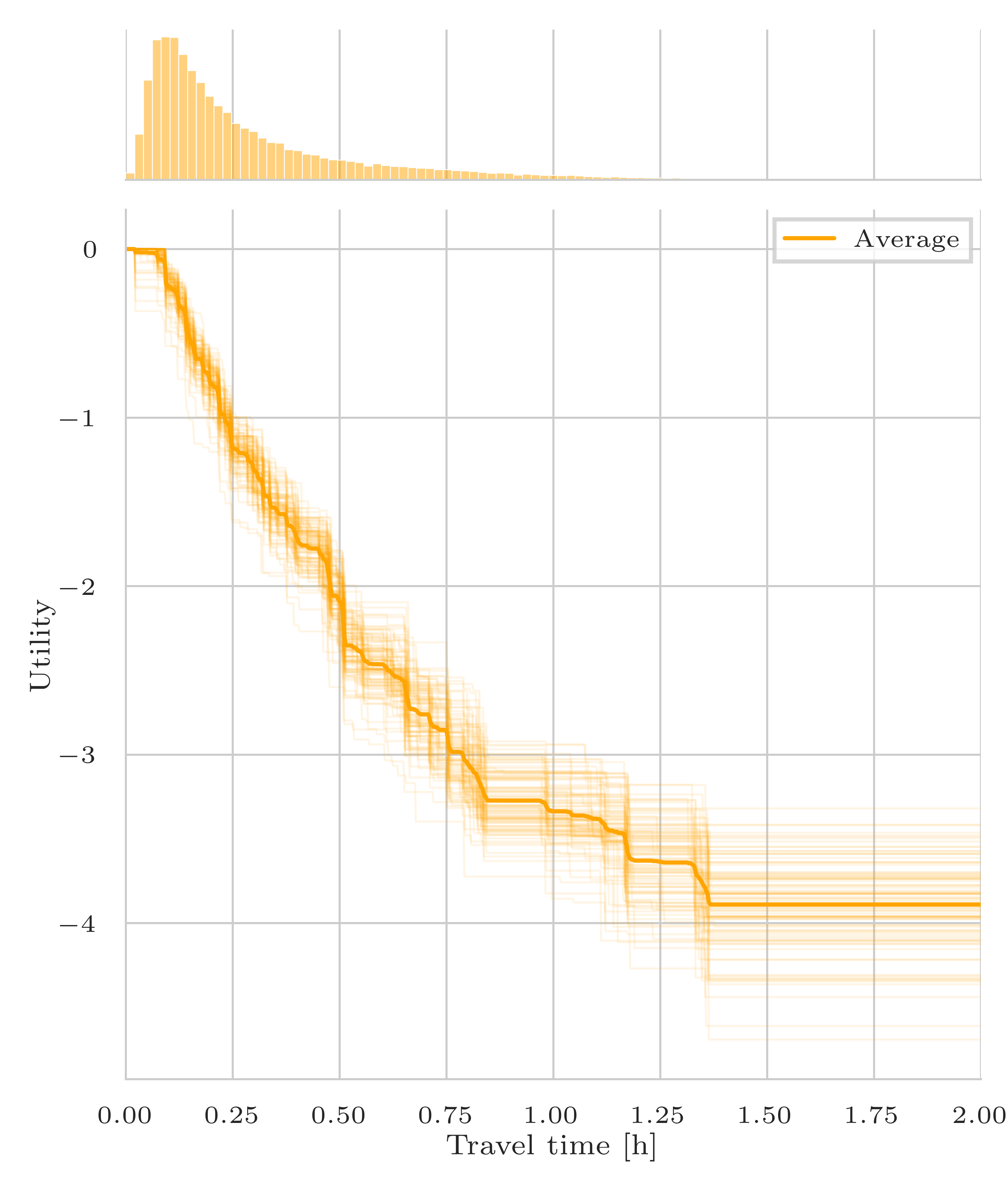}
         \caption{Driving travel time}
         \label{fig:bootstrap_driving}
     \end{subfigure}
     \caption{\centering Utility contributions of the travel time on the LPMC dataset with bootstrapping for a) walking, b) cycling, c) PT and d) driving alternative. Each line with transparency corresponds to a bootstrap sampling iteration. The mean is highlighted, and the distribution of data is shown on top of each figure. The figures are cropped at 2 hours of travel time.}
     \label{fig:bootstrap}
 \end{figure}

To demonstrate the robustness of the GBUV outputs, we perform bootstrap sampling
for 100 iterations. We plot the resulting parameter values
with their average and show the results for the travel time parameters in Figure 
\ref{fig:bootstrap}. To visualise the distribution of the input space, we plot a histogram of the data distribution on top of each figure.

Figure \ref{fig:bootstrap} show that, even with
sampling with replacement, i.e. changing the distribution of the population,
the utility values are robust. While their values may vary slightly, their
shape is similar in all utilities, especially when the density of observations
is high. The walking and driving travel time (Figures \ref{fig:bootstrap_walking} and \ref{fig:bootstrap_driving}) 
exhibit the best robustness, while cycling travel time (Figure \ref{fig:bootstrap_cycling}) and the rail travel time (Figure \ref{fig:bootstrap_PT})
are the one showing the most scale variability. However, the cycling alternative
is the least chosen alternative, and a high number of individuals
have no travel time for the PT alternative, which can explain this finding.
These bootstrapped utilities could be further used to calculate confidence interval.

\FloatBarrier

\subsection{Piece-wise cubic utility functions} \label{smoothing}

We make use of the SciPy \citep{2020SciPy-NMeth} implementation of monotonic cubic splines \citep{fritsch1980monotone, fritsch1984method} to smooth the GBUV outputs to produce piece-wise cubic utility functions. We treat finding the number and location of the knots as a heurisitic optimisation problem where we minimise the BIC of the model likelihood, following the methodology introduced in Section \ref{splines}. We make use of the Hyperopt Python library to identify optimal solutions. The case-study model has 15 variables for which we wish to extract marginal utilities. Each search in Hyperopt involves selecting a different number of knots, constrained to be an integer value between a minimum of 3 and up to 8. In total, 25 searches are conducted (i.e. 25 different combinations of numbers of knots for each variable). 

The inner optimisation loop then identifies optimal knot locations, given a fixed number of knots for each variable, using the SLSQP (Sequential Least Squares Quadratic Programming) algorithm, implemented in SciPy. We constrain the first and last knots to be at the location of the first and last observations for each variable. For the initial solution for the remaining knots, we follow \citet{wang2023monotone}, where we set the position 
of $K+1$ knots at the $k/(K+1)^{th}$ quantile.

\begin{figure}[ht]
    \centering
     \begin{subfigure}[b]{0.49\textwidth}
         \centering
         \includegraphics[width=\textwidth]{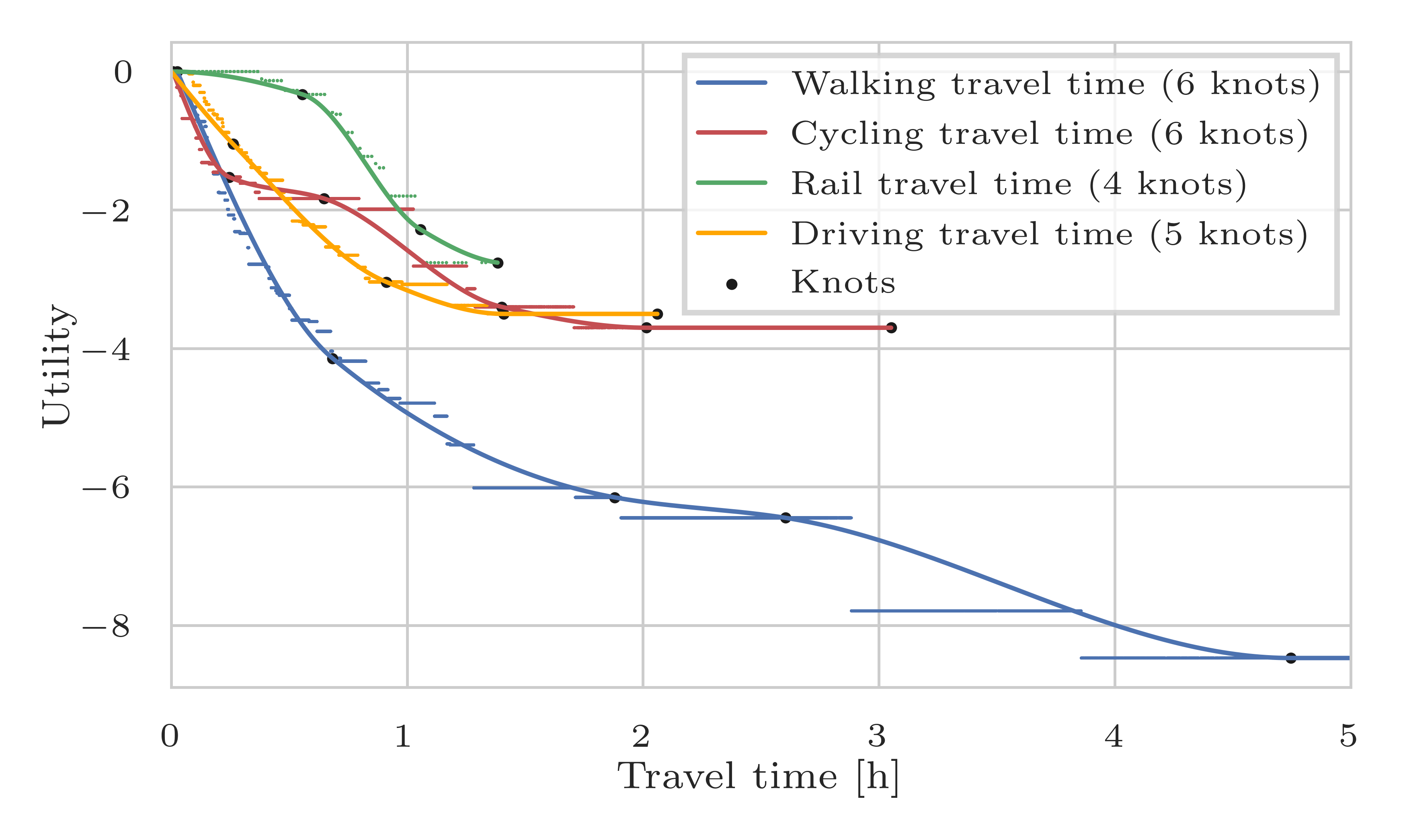}
         \caption{Travel time}
         \label{fig:lpmc_tt_splines}
     \end{subfigure}
     \begin{subfigure}[b]{0.49\textwidth}
         \centering
         \includegraphics[width=\textwidth]{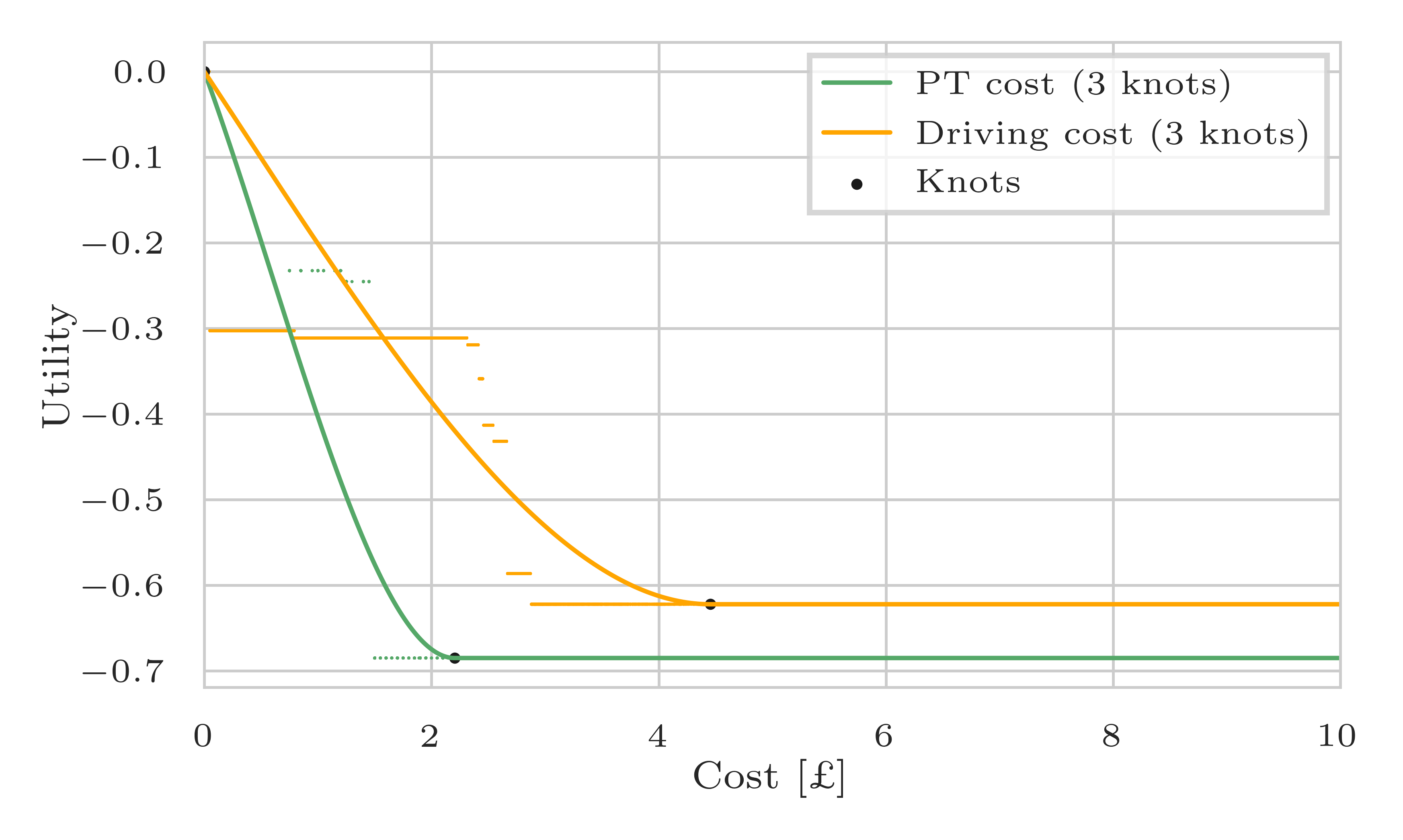}
         \caption{Cost}
         \label{fig:lpmc_cost_splines}
     \end{subfigure}
     \caption{\centering Piece-wise monotonic cubic spline interpolation of a) travel time and b) cost on the LPMC dataset. The knots are drawn in black and the first and last knots are omitted for clarity. The GBUV used for interpolation are plotted as a scatter plot.}
     \label{fig:splines}
\end{figure}

The optimised number of knots for each variable are shown in Table \ref{ce_IMPROVEMENT}. The use of the BIC as the objective function results in parsimonious solutions, where the number of splines is limited for simple transformations. The straight-line distance is not included for public transport as there
were no regression trees in the parameter ensemble. The smoothed PCUF outputs for the travel time and cost parameters are shown in Figure \ref{fig:splines}. From the graph, it is clear that PCUF output has successfully smoothed the GBUV functions, rather than overfitting with too many splines.

\begin{table}[ht]
\centering
\caption{\centering Optimal number of knots for PCUF. The number of knots is chosen with a hyperparameter search of 25 iterations}
\label{ce_IMPROVEMENT}
\begin{tabular}{lr}
\toprule
 \textbf{Attributes}                     & \textbf{Number of knots} \\ \midrule
\textit{Walking}          &                                     \\
Travel time               & 6                                   \\
Distance                  & 6                                   \\
\\
\textit{Cycling}          &                                     \\
Travel time               & 6                                   \\
Distance                  & 6                                   \\
\\
\textit{Public transport} &                                     \\
Rail travel time          & 3                                   \\
Bus travel time           & 4                                   \\
Access travel time        & 4                                   \\
Interchange waiting time  & 8                                   \\
Interchange walking time  & 7                                   \\
Cost                      & 3                                   \\
\\
\textit{Driving}          &                                     \\
Travel time               & 5                                   \\
Distance                  & 4                                   \\
Cost                      & 3                                   \\ \bottomrule
\end{tabular}
\end{table}

\begin{figure}[h!]
\centering
     \begin{subfigure}[b]{0.49\textwidth}
         \centering
         \includegraphics[width=\textwidth]{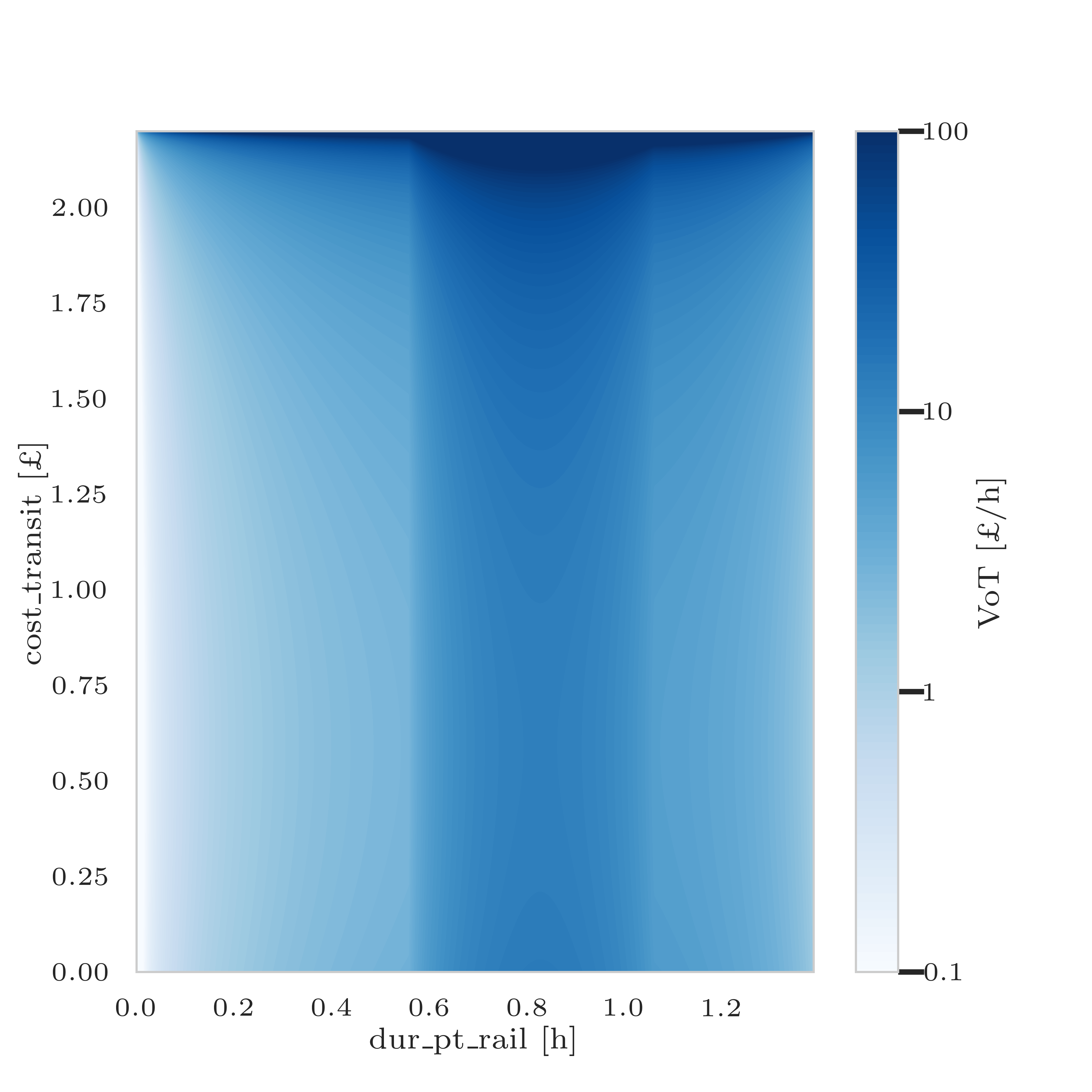}
         \caption{Rail (LPMC dataset)}
         \label{fig:vot_rail}
     \end{subfigure}
     \begin{subfigure}[b]{0.49\textwidth}
         \centering
         \includegraphics[width=\textwidth]{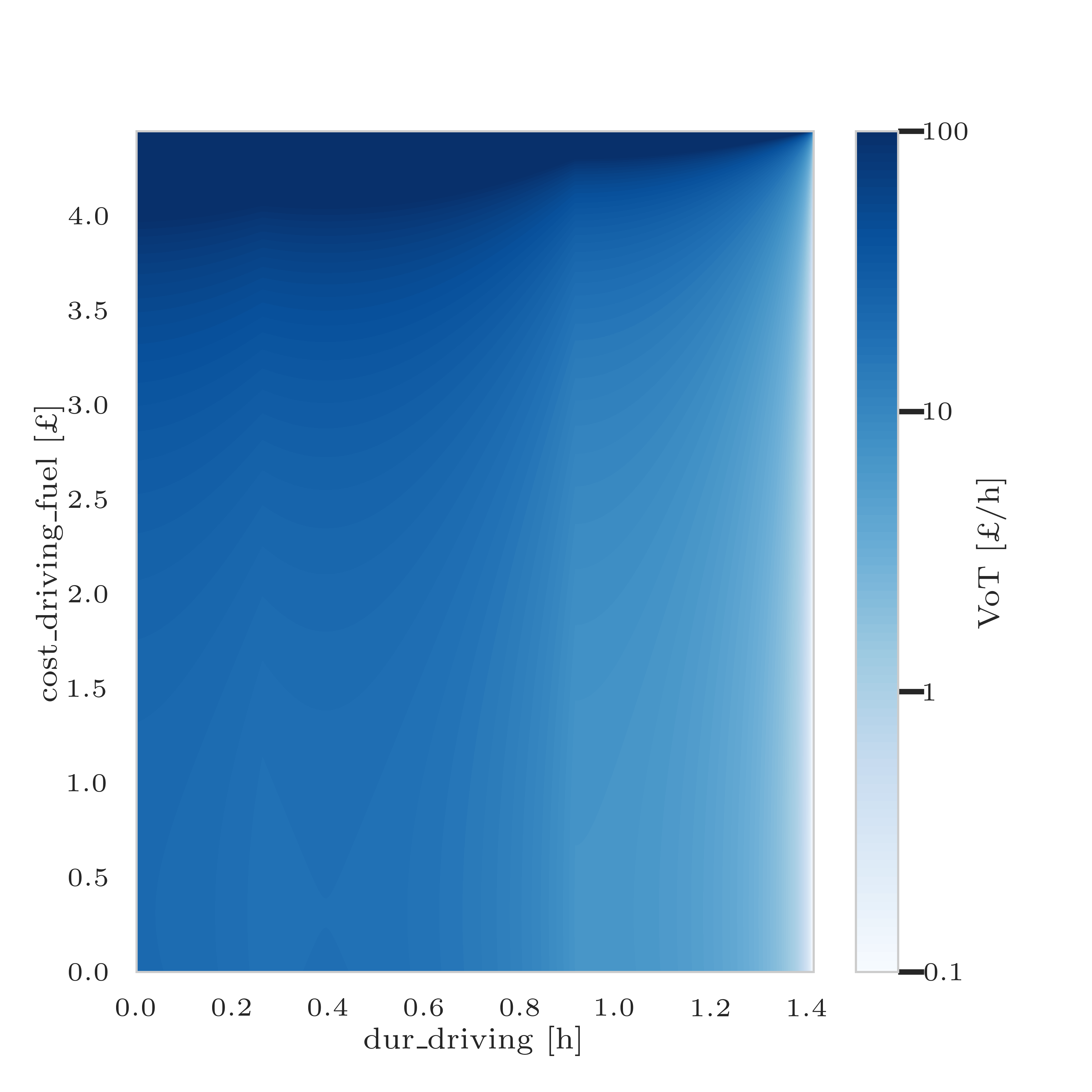}
         \caption{Driving alternative (LPMC dataset)}
         \label{fig:vot_drive}
     \end{subfigure}
        \caption{\centering Value of Time (VoT) for a) rail, b) driving. The VoT is capped at 100£/h, and displayed only where the utility functions derivatives are non zero.}
        \label{fig:vot}
\end{figure}

Using the gradients from the PCUF functions, the Value of Time (VoT) can be computed for the alternatives that include both travel 
time and cost parameters.  
We define the plot only in the area where the cost derivative is not zero, and we cap the maximum
value at 100 £/h and minimum value at 0.1 £/h. We also exclude the flat areas at low and high values of both attributes.
As both the travel time and cost attributes are non-linear, the VoT is unique to each combination of travel time and cost. As such, it is represented as a 3D distribution, though note that this distribution is homogeneous across the population. 
Figure \ref{fig:vot} displays the VoT for the PT and driving alternatives on the LPMC dataset on a logarithmic (base 10) scale. We use a logarithmic scale to better visualise the differences among lower VoTs.
The value of time of rail (Figure \ref{fig:vot_rail}) ranges from 2 to 5 £/h for trips lasting less than 0.6 hour and increases to 10 
to 20 £/h for travel times of 0.6 to 1 hour. It decreases again after one hour. This suggests that between 0.6 and 1 hour of travel time, individuals 
are willing to pay more to reduce their travel times. Regarding the value of time for driving (Figure \ref{fig:vot_drive}), 
we observe overall a decrease of the travel time with increasing travel time, and an increasing of VoT with
increasing cost. The lowest value is around 2 £/h for no cost and 1.3 hours of travel time, and the highest value 
is capped at 100 £/h for 4£ cost and no travel time. We deduce that individuals with high cost and low travel times
are willing to pay more to reduce their travel time.

\begin{figure}[h!]
\centering
     \begin{subfigure}[b]{0.49\textwidth}
         \centering
         \includegraphics[width=\textwidth]{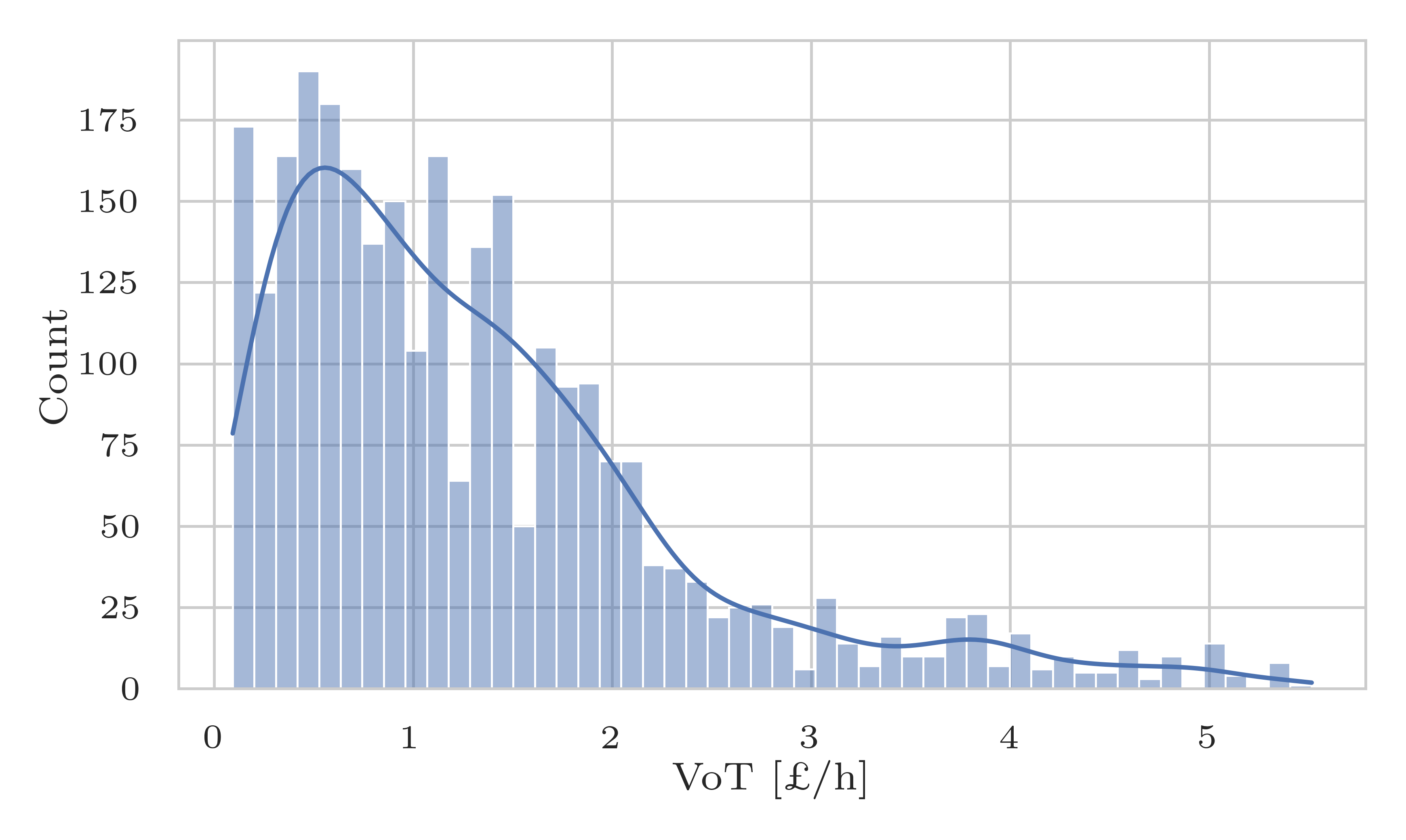}
         \caption{PT alternative}
         \label{fig:pop_vot_pt}
     \end{subfigure}
     \begin{subfigure}[b]{0.49\textwidth}
         \centering
         \includegraphics[width=\textwidth]{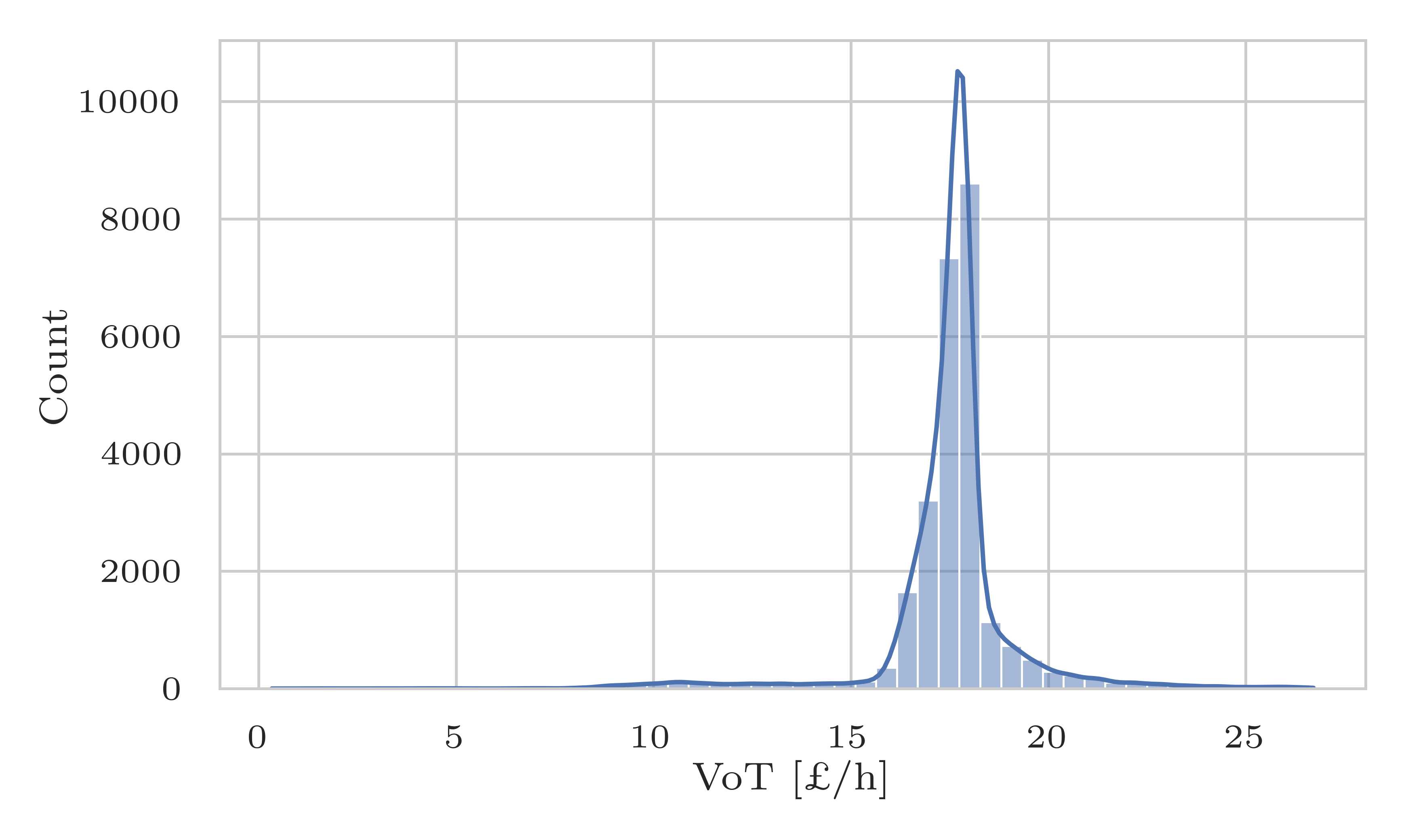}
         \caption{Driving alternative}
         \label{fig:pop_vot_drive}
     \end{subfigure}
        \caption{\centering Histogram of the population Value of Time (VoT) for a) rail, b) driving. The observations with zero travel times, as well as the highest 0.1\% VoT values are excluded. The solid line represents the kernel density estimates.}
        \label{fig:pop_vot}
\end{figure}

In addition to the contour plot VoT, we also compute the VoT across the population. To do so, we remove observations that 
have a zero travel time for the rail alternative, and exclude the 0.1\% highest values (from 99.9\% to 100\%). 
Then, we calculate the VoT of all remaining individuals with their respective costs and travel times.
The results are shown in Figure \ref{fig:pop_vot}. 
For the PT alternative (Figure \ref{fig:pop_vot_pt}), the distribution of VoT peaks below 1 £/h and then continuously decreases until 5£/h. For driving VoT (Figure \ref{fig:pop_vot_drive}), there is a sharp peak at 17.5 £/h. These values are both lower than the VoT extracted from linear-in-parameter RUM models for the same dataset, of 8.73 £/h and 40 £/h respectively (see \citet{hillel2019understanding}, p.133), showing the impact of the non-linear utility specification.

\section{Extensions of RUMBoost} \label{model extensions}

In this section, we present three extensions of RUMBoost, highlighting how the approach can be generalised to different modelling scenarios: a) incorporating attribute interactions (Section \ref{feature interaction}); b) assuming alternative correlation within the error term (Section \ref{nested rumboost}); and c) accounting for
correlation within trips made by the same individual (Section \ref{funct effect rumboost}).

\subsection{Second-order attribute interactions} \label{feature interaction} 

\begin{figure}[ht!]
    \centering
     \begin{subfigure}[b]{0.49\textwidth}
         \centering
         \includegraphics[width=\textwidth]{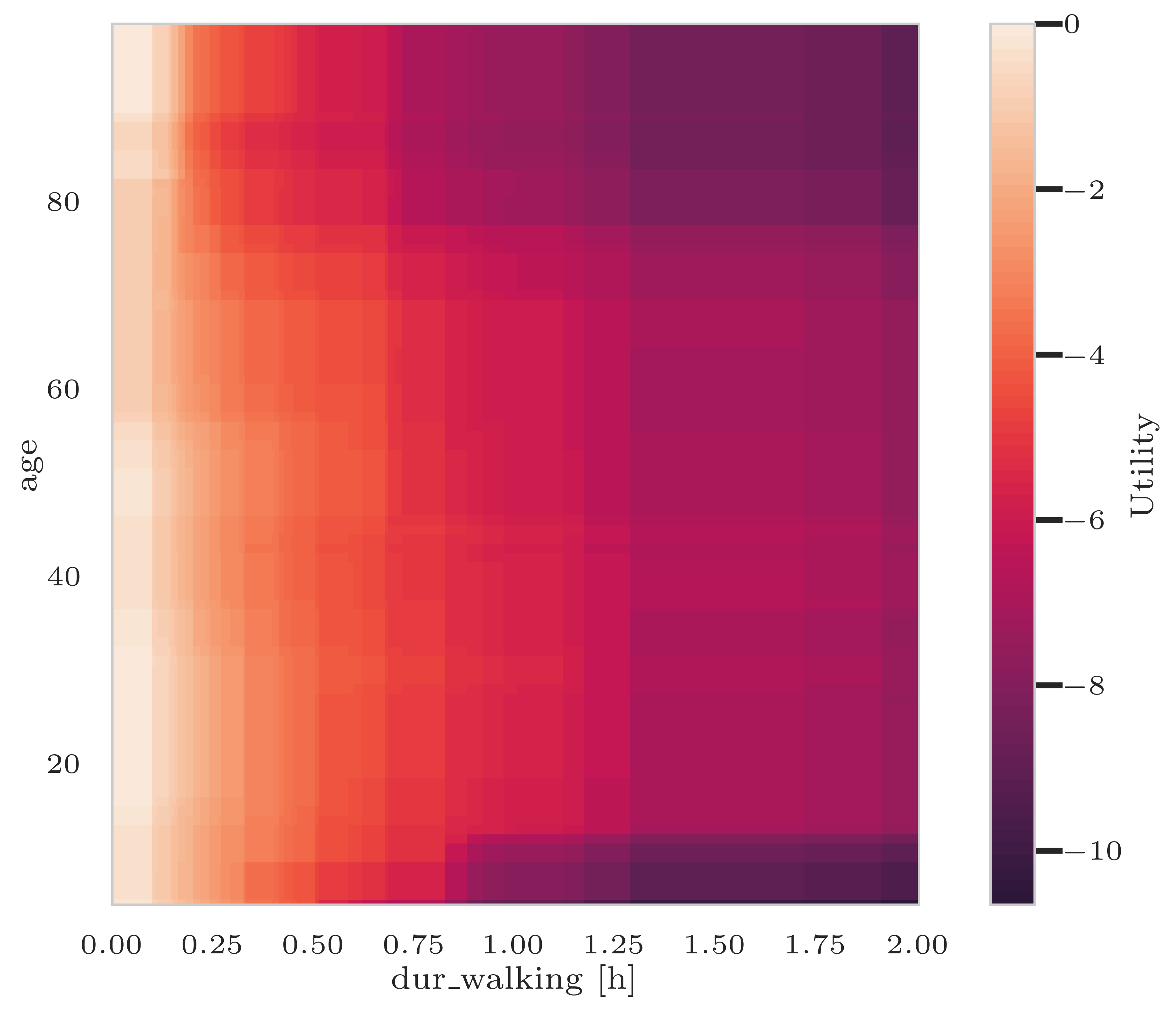}
         \caption{Walking alternative}
         \label{fig:walk_tt_age}
     \end{subfigure}
     \hfill
     \begin{subfigure}[b]{0.49\textwidth}
         \centering
         \includegraphics[width=\textwidth]{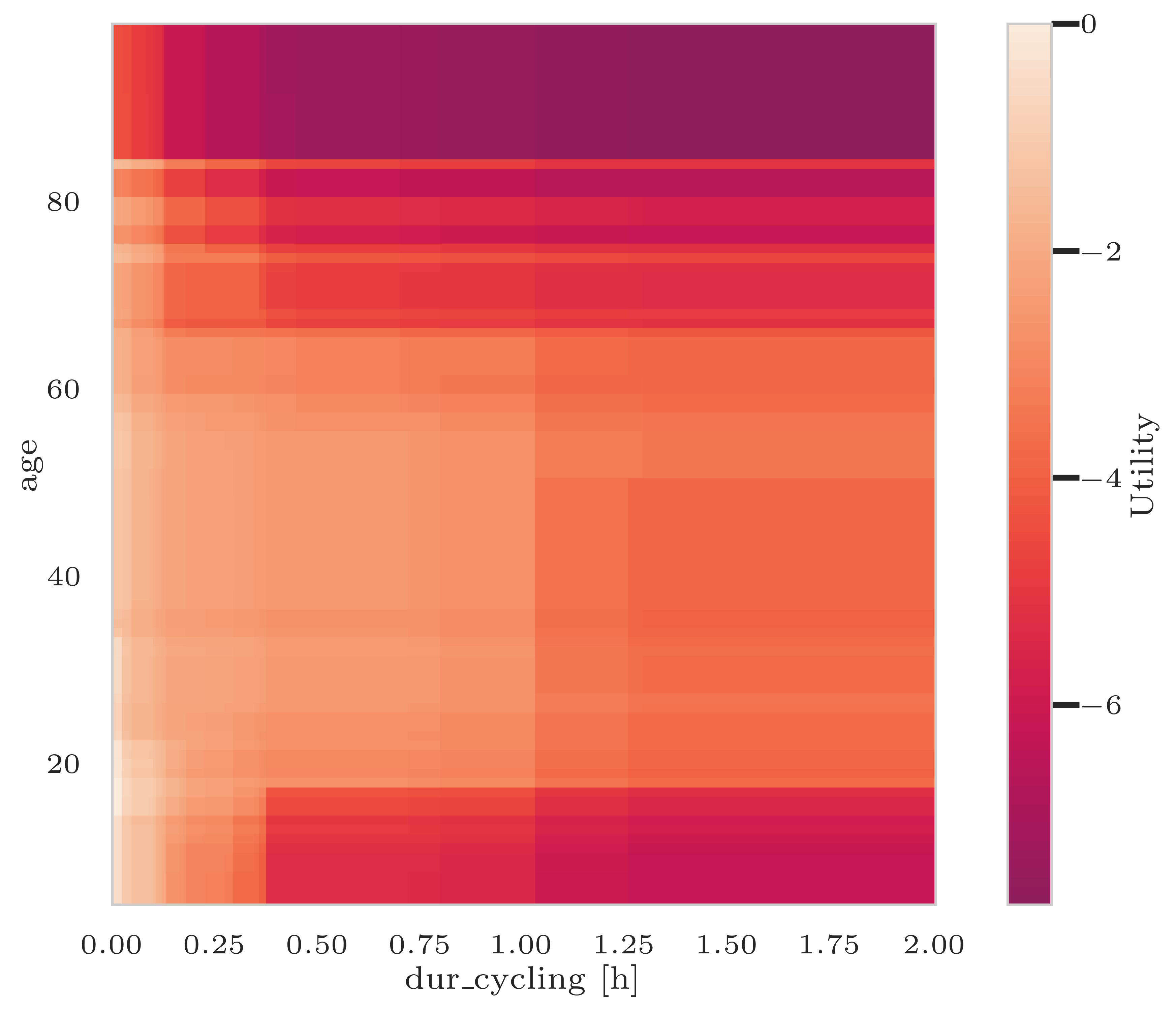}
         \caption{Cycling alternative}
         \label{fig:cycle_tt_age}
     \end{subfigure}
     \begin{subfigure}[b]{0.49\textwidth}
         \centering
         \includegraphics[width=\textwidth]{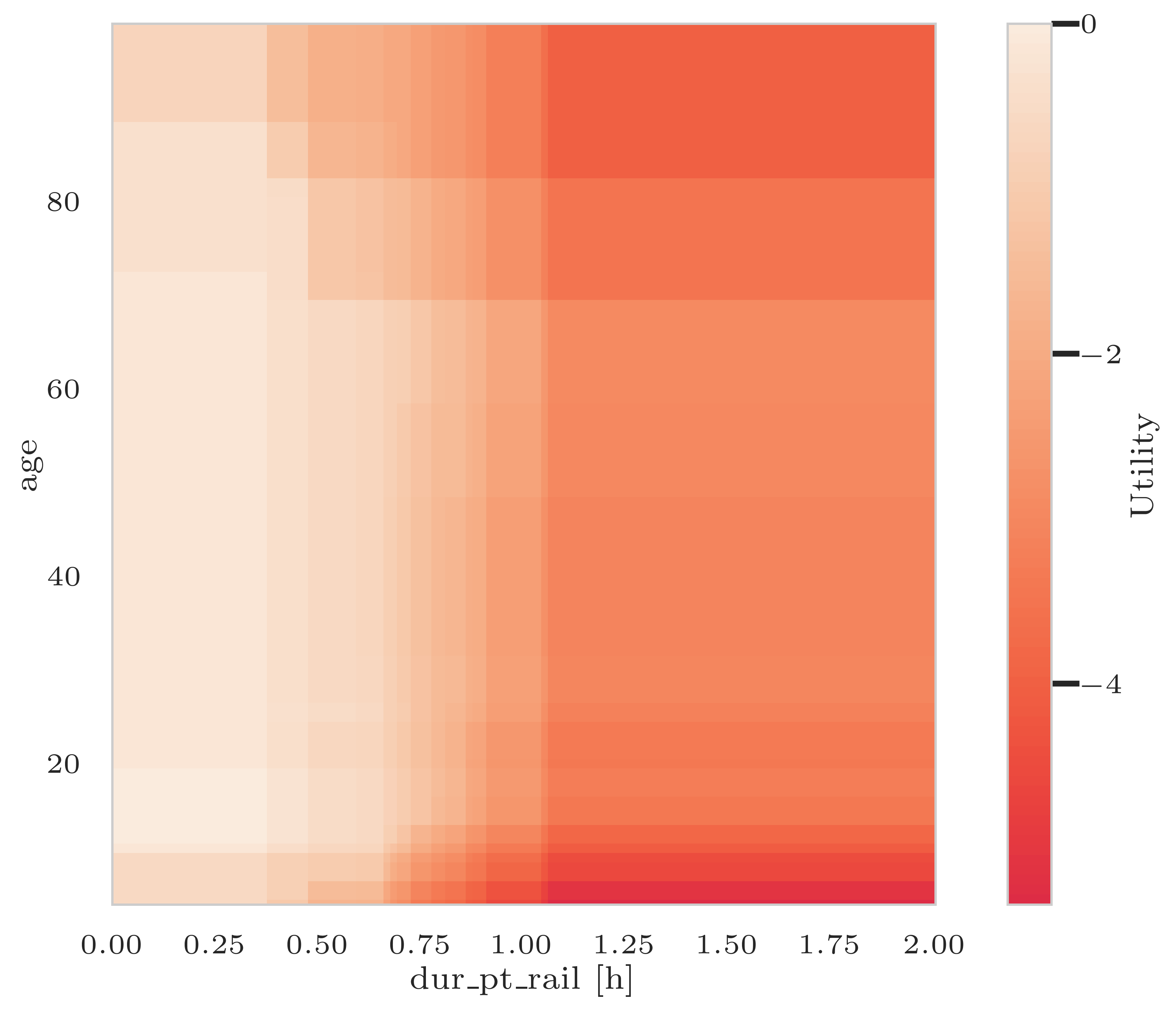}
         \caption{Public transport alternative}
         \label{fig:pt_tt_age}
     \end{subfigure}
     \begin{subfigure}[b]{0.49\textwidth}
         \centering
         \includegraphics[width=\textwidth]{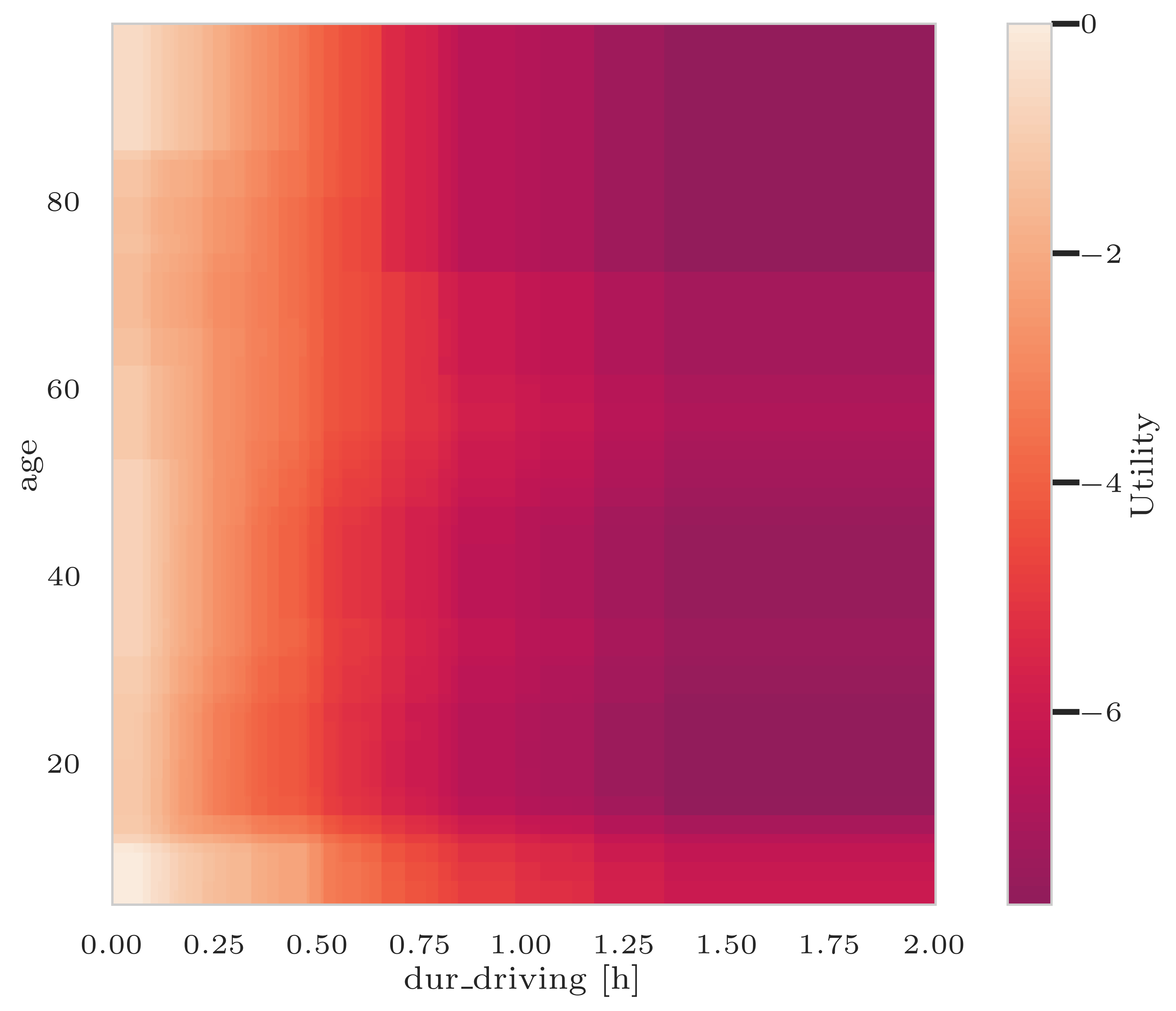}
         \caption{Driving alternative}
         \label{fig:drive_tt_age}
     \end{subfigure}
        \caption{\centering Utility function in the form of a contour plot for the travel time (x axis) with age (y axis) interaction on the LPMC dataset for a) walking, b) cycling, c) PT and d) driving alternative. Only the travel time is subject to monotonicity constraint.}
        \label{fig:2d}
\end{figure}

By allowing two continuous attributes to interact, we can consider 
attribute interactions. In doing so, it is still possible to interpret the ensemble 
output for these two attributes on a contour plot. As an example, we arbitrarily allow age and travel time 
to interact. We allow for a max depth of two in each tree to allow for feature interactions and, through early stopping, perform 680 total boosting rounds. 
Figure \ref{fig:2d} shows the resulting contour plot for all alternatives. The contour plots are
only monotonic with respect to travel time, as specified in the model. For the walking alternative
(Figure \ref{fig:walk_tt_age}), longer travel times lead to increased disutility for both younger 
and older ages, while shorter travel times result in relatively uniform disutility. For cycling 
(Figure \ref{fig:cycle_tt_age}), we observe a similar phenomenon for longer travel times
but the disutility is still pronounced for older individuals with shorter travel times, 
which aligns with the expectation that cycling may be less feasible for older individuals.
In the case of public transport (Figure \ref{fig:pt_tt_age}), disutility is more significant 
for younger ages and less pronounced for adults and shorter travel times. Lastly, the disutility
associated with an increase in age or travel time for the car alternative (Figure \ref{fig:drive_tt_age}) remains relatively 
constant, but it is mitigated for ages below 10 and travel times shorter than 0.5 hour.

\subsection{Nested RUMBoost}\label{nested rumboost}

Until now, we assumed the error terms for each alternative to be distributed i.i.d., leading to the MNL model formulation. We show that we can
relax this assumption with our approach by assuming an error term correlated within alternatives.
In RUM, this error term leads to the Nested Logit (NL) model. We update the probability formula and update the gradient and hessian (second derivative) computations
accordingly, giving:

\begin{equation}
    P(i) = P(i|m)P(m)
\end{equation}
where the probability of choosing $i$ knowing the nest $m$ is:

\begin{equation}
    P(i|m) = \frac{e^{\mu_m V_{i}}}{\sum_{j\in m}e^{\mu_m V_{j}}},
\end{equation}
while the probability to choose the nest $m$ is:

\begin{equation}
    P(m) = \frac{e^{\Tilde{V}_{m}}}{\sum_{p=1}^{M}e^{\Tilde{V}_{p}}},
\end{equation}
where:
\begin{itemize}
    \item $\Tilde{V}_{m} = \frac{1}{\mu_m} \ln\left( \sum_{i \in m} e^{\mu_m V_{i}} \right)$
    \item $M$ the number of nest
    \item $\mu_m$ the scaling parameter of nest $m$
\end{itemize}

We treat $\mu$ as a hyperparameter,
where we search its optimal value with a 5-fold cross validation scheme on the training test.
More details are given in the appendix \ref{app:hyperparamers}. We train the Nested RUMBoost model on
the LPMC dataset, and compare its performance against a NL model. The nest is assumed to be
within the motorised modes (PT and driving) while walking and cycling are in their own nests.
The optimal value of $\mu$ found with hyperparameter tuning is $1.16$, while the one estimated 
with the Nested Logit model is $1.35$ (see \ref{app:mnl}). The difference of flexibility of these
two models explains this difference. The results are shown in Table \ref{tab:nested}. 

\begin{table}[ht]
\centering
\caption{\centering Benchmark of classification on the LPMC Dataset for the Nested RUMBoost with the RUMs and RUMBoost. The models are compared with their CEL (negative Cross-Entropy Loss, lower the
better) on the test set and their computational time for one CV iteration.}
\begin{tabular}{lcrcrcrcr} \toprule
                       \multicolumn{1}{c}{} & \multicolumn{2}{c}{\textbf{MNL}}                              & \multicolumn{2}{c}{\textbf{Nested Logit}}                     & \multicolumn{2}{c}{\textbf{RUMBoost-GBUV}}                         & \multicolumn{2}{c}{\textbf{Nested RUMBoost}}                  \\ \midrule
                       \multicolumn{1}{c}{} & CE                         & \multicolumn{1}{c}{Time {[}s{]}} & CE                         & \multicolumn{1}{c}{Time {[}s{]}} & CE                         & \multicolumn{1}{c}{Time {[}s{]}} & CE                         & \multicolumn{1}{c}{Time {[}s{]}} \\ \midrule
 5 fold CV         & \multicolumn{1}{r}{0.6913} & 242.14                           & \multicolumn{1}{r}{0.6921} & 1067.04                          & \multicolumn{1}{r}{0.6570} & 8.9                              & \multicolumn{1}{r}{0.6568} & 48.53                            \\
                       Holdout test set                 & \multicolumn{1}{r}{0.7085} & -                                & \multicolumn{1}{r}{0.7091} & -                                & \multicolumn{1}{r}{0.6737} & -                                & \multicolumn{1}{r}{0.6731} & -                                \\ \bottomrule
\end{tabular}
\label{tab:nested}
\end{table}

Whilst the Nested Logit model does not show a performance improvement over the MNL mode in terms of out-of-sample validation, the nested RUMBoost model marginally outperforms the base RUMBoost model. Note that these results are without PCUF smoothing, but the 
Nested RUMBoost could be smoothed, as in Section \ref{smoothing}. 

\subsection{Functional Effect RUMBoost}\label{funct effect rumboost}

Lastly, we propose the Functional Effect RUMBoost (FE-RUMBoost), a model accounting for observations
correlation. This model draws some parallels with the Mixed Effect model, where the
fixed effect part is RUMBoost without attribute interaction and the random
effect part includes all socio-economic characteristics with full interaction. By doing so,
we keep the full utility interpretability on the trip attributes, and we learn an individual-
specific constant for each alternative with the second part of the model. In other words, the second part of the model
is a latent representation of each individual from their socio-economic characteristics, 
which can capture correlation for panel data or other observation correlated situations. 
We apply the model on the LPMC dataset and we compare it with the benchmarks of Section \ref{benchmarks}
in Table \ref{tab:fe_model}.

\begin{table}[ht]
\centering
\caption{\centering Benchmark of classification on the LPMC Dataset for the FE-RUMBoost with the ML classifiers. The models are compared with their CEL (negative Cross-Entropy Loss, lower the
better) on the test set and their computational time for one CV iteration.}
\label{tab:fe_model}
\begin{tabular}{lcrcrcrcr} \toprule
                       \multicolumn{1}{c}{} & \multicolumn{2}{c}{\textbf{LightGBM}}                         & \multicolumn{2}{c}{\textbf{NN}}                               & \multicolumn{2}{c}{\textbf{DNN}}                              & \multicolumn{2}{c}{\textbf{FE-RUMBoost}}                      \\ \midrule
                       \multicolumn{1}{c}{} & CE                         & \multicolumn{1}{c}{Time {[}s{]}} & CE                         & \multicolumn{1}{c}{Time {[}s{]}} & CE                         & \multicolumn{1}{c}{Time {[}s{]}} & CE                         & \multicolumn{1}{c}{Time {[}s{]}} \\ \midrule
  5 fold CV         & \multicolumn{1}{r}{0.6381} & 4.64                             & \multicolumn{1}{r}{0.6516} & 7.85                             & \multicolumn{1}{r}{0.6613} & 3.89                             & \multicolumn{1}{r}{0.6447} & 10.9                             \\
                       Holdout test set                 & \multicolumn{1}{r}{0.6537} & -                                & \multicolumn{1}{r}{0.6667} & -                                & \multicolumn{1}{r}{0.6735} & -                                & \multicolumn{1}{r}{0.6626} & -                                \\ \bottomrule
\end{tabular}
\end{table}

\begin{figure}[ht!]
    \centering
    \includegraphics[width=\linewidth]{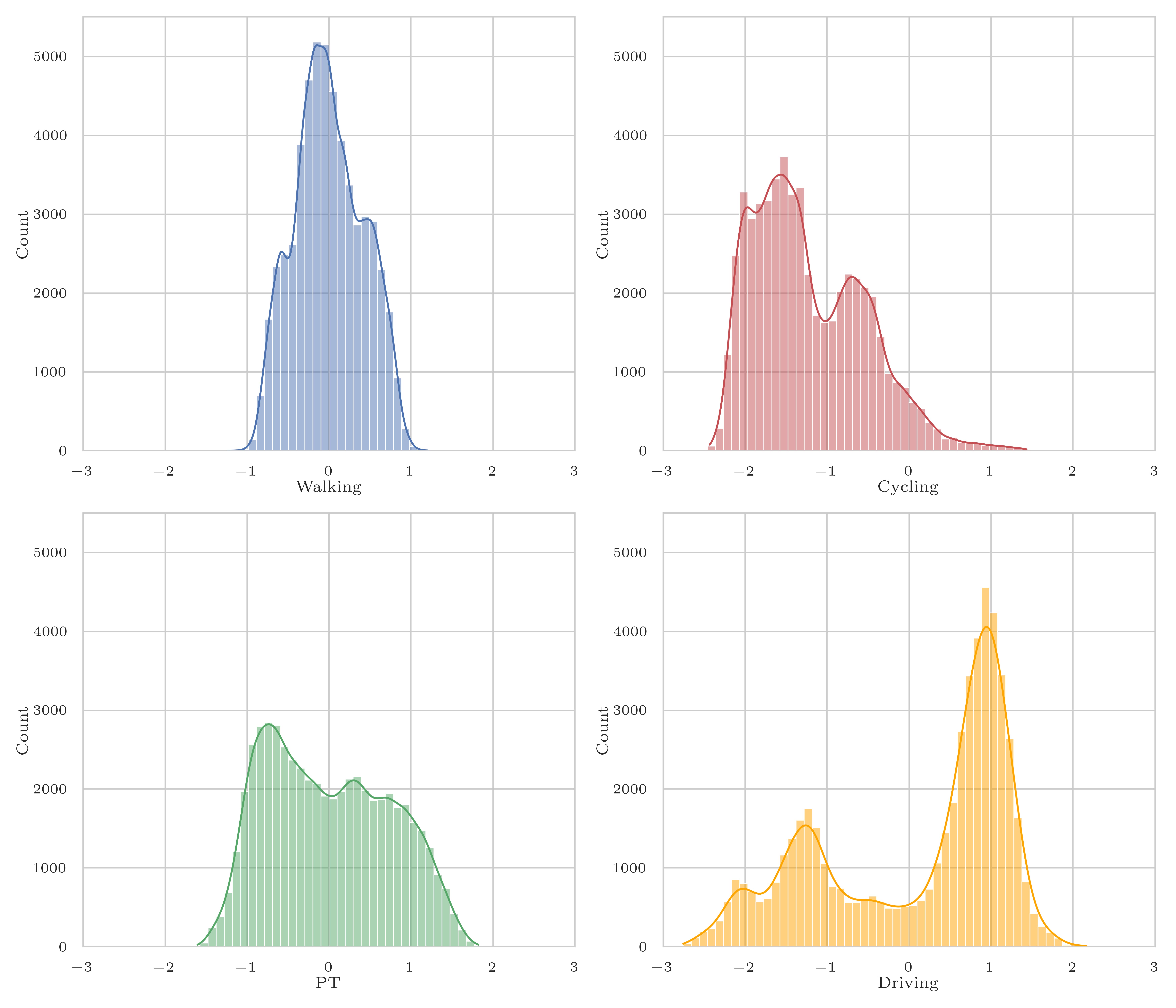}
    \caption{\centering Histograms of the individual-specific constant learnt with the Functional Effect RUMBoost of a) walking, b) cycling, c) PT and d) driving alternative. For each histogram, we plot the kernel density estimate as a solid line. Overall, cycling, PT and driving individual constants exhibit a bi-modal distribution, and walking alternative a uni-modal distribution centered at zero.}
    \label{fig:ind_spec}
\end{figure}

This extension of the model substantially improves the prediction performance on the test set.
Our model outperforms both the NN and DNN classifiers and narrows the gap vs the unconstrained LightGBM model, while keeping the full interpretability
on key alternative-specific attributes. The computational cost induced by the greater complexity is minimal 
compared to the initial RUMBoost. Again, the model presented here is without smoothing but,
because all the trip attributes that were previously smoothed are in the first part of the model,
we can apply smoothing as in Section \ref{smoothing}. In addition, we can 
visualise the individual specific constant per alternative. We show the distribution of the constants
per alternative in Figure \ref{fig:ind_spec} in the form of histograms. These histograms show clearly that for cycling, PT and driving the distribution of individual-specific
constants are bi-modal.

These three extensions demonstrate the RUMBoost model's ability to incorporate attribute interactions, account for correlated error terms within alternatives, and learn an individual-specific constant for observation correlated data. The extensions can also be combined, as they are applied to different parts of the model. This showcases the flexibility and generalisability of the RUMBoost framework. In future work, any complex choice situations that lead to a defined gradient and hessian could be applied to the model. 

\section{Conclusion and further work}
\label{Conclusion and further work}

The methodology presented in this paper allows for a fully interpretable ML model 
(RUMBoost) based on GBDT and inspired by random utility models. In short, RUMBoost
replaces every parameter of a RUM by an ensemble of regression trees. By re-implementing 
classification for GBDT, we can provide specific attributes for alternative utilities, 
control attribute interactions in the ensemble, and apply monotonic constraints on 
key attributes based on domain knowledge. These constraints considerably improve the 
predictions of traditional RUMs and enable the derivation of non-linear utility 
functions. Furthermore, we apply piece-wise monotonic cubic splines to interpolate the utility 
functions and obtain a smooth utility function. We find that the smoothing acts as further
regularization and enables us to compute the Value of Time (VoT). We also show that the
modularity of our approach allow for the estimation of complex model specification such
as error term accounting for correlation within alternatives or correlation within observations. Our approach offer
to observe the full functional form of the utility function with defined gradient, just like in DCMs. The 
key difference is that the utility function is directly learnt from the data.

Whilst applied here to choice models, this methodology could be used in place of any linear-in-parameters models, for regression, classification, or any task for which the gradient and hessian of the cost function are well defined. Further work 
includes applying the model to various problems to demonstrate this statement.
Further work also includes applying the RUMBoost model to other complex model specifications
such as the Cross-Nested Logit (CNL) model. The PCUF algorithm could be improved
by applying B-splines, which would provide a $C^2$ monotonic interpolation of the data,
where shape constraint could be incorporated. Finally, the GBUV could be computed
directly with linear trees, quadratic trees or splines, to obtain directly 
piece-wise utility functions with defined gradient.

\section*{CRediT authorship contribution statement} \label{credit}
\textbf{Nicolas Salvadé}: Conceptualization, Methodology, Software, Writing - Original Draft, Visualization. \textbf{Tim Hillel}: Conceptualization, Methodology, Writing - Review and Editing, Supervision.

\section*{Declaration of competing interest} \label{interest}

None

\section*{Acknowledgements} \label{acknowledgements}

We sincerely thank Prof. Michel Bierlaire for his guidance on the smoothing process. We also would like to express our sincere gratitude to Dr. José Ángel Martín-Baos for his assistance with the benchmarks on the LPMC.

\appendix

\section{Hyperparameter search}
\label{app:hyperparamers}

\begin{table}[H]
\centering
\caption{\centering Hyperparameter search and optimal value for RUMBoost, Nested RUMBoost, and FE-RUMBoost on the LPMC dataset}
\begin{tabular}{llrrrr} \toprule
\multicolumn{3}{l}{}                                            & \textbf{RUMBoost} & \textbf{Nested RUMBoost} & \textbf{FE-RUMBoost} \\ \midrule
\multicolumn{3}{l}{Number of searches}                            & 1        & 25              & 100         \\
\multicolumn{3}{l}{Time {[}s{]}}                                & 44.5     & 5378            & 5366        \\ \midrule
\textit{Hyperparameter}     & \textit{Distribution}   & \textit{Search space}     &          &                 &             \\
Mean of num\_iterations  & early stopping & 100              & 1300     & 1256            & 1099        \\
bagging\_fraction           & uniform        & {[}0.5, 1{]}     & -        & -               & 0.700       \\
bagging\_freq               & choice         & (0, 1, 5, 10)    & -        & -               & 10          \\
feature\_fraction           & uniform        & {[}0.5, 10{]}    & -        & -               & 0.867       \\
lambda\_l1                  & log uniform    & {[}0.0001, 10{]} & -        & -               & 6.592       \\
lambda\_l2                  & log uniform    & {[}0.0001, 10{]} & -        & -               & 1.028       \\
learning\_rate              & fixed          & 0.1              & -        & -               & 0.1         \\
max\_bin                    & uniform        & {[}100, 500{]}   & -        & -               & 131         \\
min\_data\_in\_leaf         & uniform        & {[}1, 200{]}     & -        & -               & 27          \\
min\_gain\_to\_split        & log uniform    & {[}0.0001, 5{]}  & -        & -               & 0.800       \\
min\_sum\_hessian\_in\_leaf & log uniform    & {[}1,100{]}      & -        & -               & 1.783       \\
num\_leaves                 & uniform        & {[}2, 100{]}     & -        & -               & 74          \\
mu                          & uniform        & {[}1, 2{]}       & -        & 1.167           & -            \\ \bottomrule
\end{tabular}
\end{table}

\begin{table}[H]
\centering
\caption{\centering Hyperparameter search and optimal value for LightGBM, NN, and DNN on the LPMC dataset.}
\begin{tabular}{llllll} \toprule
\multicolumn{3}{l}{}                                            & \textbf{LightGBM} & \textbf{NN}              & \textbf{DNN}        \\ \midrule
\multicolumn{3}{l}{Number of searches}                            & 1000     & 1000            & 1000        \\
\multicolumn{3}{l}{Time {[}s{]}}                                & 24882     & 28736            & 26343        \\ \midrule
\textit{Hyperparameter}     & \textit{Distribution}   & \textit{Search space}     &          &                 &             \\
Mean of CV num\_iterations  & early stopping & 100              & 1493    & -           & -        \\
bagging\_fraction           & uniform        & {[}0.5, 1{]}     & 0.7204        & -               & 0.700       \\
bagging\_freq               & choice         & (0, 1, 5, 10)    & 1        & -               & 10          \\
feature\_fraction           & uniform        & {[}0.5, 10{]}    & 0.6007        & -               & 0.867       \\
lambda\_l1                  & log uniform    & {[}0.0001, 10{]} & 0.0242        & -               & 6.592       \\
lambda\_l2                  & log uniform    & {[}0.0001, 10{]} & 0.0001        & -               & 1.028       \\
learning\_rate              & fixed          & -             & 0.1        & adaptive               & 0.1         \\
max\_bin                    & discrete uniform        & {[}100, 500{]}   & 237        & -               & 131         \\
min\_data\_in\_leaf         & discrete uniform        & {[}1, 200{]}     & 156        & -               & 27          \\
min\_gain\_to\_split        & log uniform    & {[}0.0001, 5{]}  & 0.0007        & -               & 0.800       \\
min\_sum\_hessian\_in\_leaf & log uniform    & {[}1,100{]}      & 1.4136        & -               & 1.783       \\
num\_leaves                 & discrete uniform        & {[}2, 100{]}     & 16        & -               & 74          \\
activation                          & fixed        & -       & -        & $tanh(x)$           &    $relu(x)$        \\
batch\_size                  & choice    & {(}128, 256, 512, 1024{)} & -        & 1024               & 1024      \\
hidden\_layer\_size                  & discrete uniform    & {[}10, 500{]} & -        & -               & 20       \\
learning\_rate\_init              & uniform          & {[}0.0001, 1{]}             & -        & 0.0907               & -        \\
solver         & choice        & {[}lbfgs, sgd, adam{]}     & -        & sgd               & -          \\
depth & discrete uniform    & {[}2,10{]}      & -        & -               & 2       \\
drop                 & choice        & {(}0.5, 0.3, 0.1{)}     & -        & -               & 0.3          \\
epochs                  & discrete uniform    & {[}50, 200{]} & -        & -               & 90       \\
width         & choice        & {(}25, 50, 100, 150, 200{)}     & -        & -               & 50      \\ \bottomrule
\end{tabular}
\end{table}

\section{Estimation of the MNL models}
\label{app:mnl}

\begin{center}
\begin{longtable}{lrrr}
\caption{\centering Parameter estimates of the LPMC MNL. Out of the 62 parameters, 9 are not significant at a 95\% confidence interval.} \\ \toprule
\multicolumn{4}{c}{\textbf{LPMC - MNL}}                                            \\ \midrule
                                            & Value  & Active bound & Rob. p-value \\ \midrule
                                            \endhead
ASC\_Bike                                   & -3.380 & 0.000        & 0.000        \\
ASC\_Car                                    & -2.592 & 0.000        & 0.000        \\
ASC\_Public\_Transport                      & -1.908 & 0.000        & 0.000        \\
B\_age\_Bike                                & -0.004 & 0.000        & 0.032        \\
B\_age\_Car                                 & 0.005  & 0.000        & 0.000        \\
B\_age\_Public\_Transport                   & 0.011  & 0.000        & 0.000        \\
B\_car\_ownership\_Bike                     & 0.036  & 0.000        & 0.590        \\
B\_car\_ownership\_Car                      & 0.694  & 0.000        & 0.000        \\
B\_car\_ownership\_Public\_Transport        & -0.213 & 0.000        & 0.000        \\
B\_con\_charge\_Car                         & -1.147 & 0.000        & 0.000        \\
B\_cost\_driving\_fuel\_Car                 & 0.000  & 1.000        & 1.000        \\
B\_cost\_transit\_Public\_Transport         & -0.115 & 0.000        & 0.000        \\
B\_day\_of\_week\_Bike                      & -0.020 & 0.000        & 0.201        \\
B\_day\_of\_week\_Car                       & 0.030  & 0.000        & 0.000        \\
B\_day\_of\_week\_Public\_Transport         & -0.044 & 0.000        & 0.000        \\
B\_distance\_Bike                           & -0.232 & 0.000        & 0.040        \\
B\_distance\_Car                            & 0.000  & 1.000        & 1.000        \\
B\_distance\_Public\_Transport              & 0.000  & 1.000        & 1.000        \\
B\_driving\_license\_Bike                   & 0.678  & 0.000        & 0.000        \\
B\_driving\_license\_Car                    & 0.663  & 0.000        & 0.000        \\
B\_driving\_license\_Public\_Transport      & -0.526 & 0.000        & 0.000        \\
B\_dur\_cycling\_Bike                       & -2.670 & 0.000        & 0.000        \\
B\_dur\_driving\_Car                        & -4.777 & 0.000        & 0.000        \\
B\_dur\_pt\_access\_Public\_Transport       & -4.608 & 0.000        & 0.000        \\
B\_dur\_pt\_bus\_Public\_Transport          & -1.911 & 0.000        & 0.000        \\
B\_dur\_pt\_int\_waiting\_Public\_Transport & -4.284 & 0.000        & 0.000        \\
B\_dur\_pt\_int\_walking\_Public\_Transport & -2.335 & 0.000        & 0.027        \\
B\_dur\_pt\_rail\_Public\_Transport         & -1.338 & 0.000        & 0.000        \\
B\_dur\_walking\_Walk                       & -8.596 & 0.000        & 0.000        \\
B\_female\_Bike                             & -0.834 & 0.000        & 0.000        \\
B\_female\_Car                              & 0.100  & 0.000        & 0.002        \\
B\_female\_Public\_Transport                & 0.160  & 0.000        & 0.000        \\
B\_fueltype\_Avrg\_Bike                     & -0.691 & 0.000        & 0.000        \\
B\_fueltype\_Avrg\_Car                      & -1.400 & 0.000        & 0.000        \\
B\_fueltype\_Avrg\_Public\_Transport        & -0.221 & 0.000        & 0.000        \\
B\_fueltype\_Diesel\_Bike                   & -0.822 & 0.000        & 0.000        \\
B\_fueltype\_Diesel\_Car                    & -0.228 & 0.000        & 0.000        \\
B\_fueltype\_Diesel\_Public\_Transport      & -0.419 & 0.000        & 0.000        \\
B\_fueltype\_Hybrid\_Bike                   & -1.000 & 0.000        & 0.000        \\
B\_fueltype\_Hybrid\_Car                    & -0.721 & 0.000        & 0.000        \\
B\_fueltype\_Hybrid\_Public\_Transport      & -0.945 & 0.000        & 0.000        \\
B\_fueltype\_Petrol\_Bike                   & -0.867 & 0.000        & 0.000        \\
B\_fueltype\_Petrol\_Car                    & -0.242 & 0.000        & 0.000        \\
B\_fueltype\_Petrol\_Public\_Transport      & -0.323 & 0.000        & 0.000        \\
B\_pt\_n\_interchanges\_Public\_Transport   & -0.101 & 0.000        & 0.154        \\
B\_purpose\_B\_Bike                         & -0.029 & 0.000        & 0.775        \\
B\_purpose\_B\_Car                          & -0.043 & 0.000        & 0.543        \\
B\_purpose\_B\_Public\_Transport            & -0.012 & 0.000        & 0.874        \\
B\_purpose\_HBE\_Bike                       & -1.054 & 0.000        & 0.000        \\
B\_purpose\_HBE\_Car                        & -0.756 & 0.000        & 0.000        \\
B\_purpose\_HBE\_Public\_Transport          & -0.237 & 0.000        & 0.000        \\
B\_purpose\_HBO\_Bike                       & -0.773 & 0.000        & 0.000        \\
B\_purpose\_HBO\_Car                        & -0.352 & 0.000        & 0.000        \\
B\_purpose\_HBO\_Public\_Transport          & -0.442 & 0.000        & 0.000        \\
B\_purpose\_HBW\_Bike                       & -0.291 & 0.000        & 0.000        \\
B\_purpose\_HBW\_Car                        & -1.062 & 0.000        & 0.000        \\
B\_purpose\_HBW\_Public\_Transport          & -0.502 & 0.000        & 0.000        \\
B\_purpose\_NHBO\_Bike                      & -1.233 & 0.000        & 0.000        \\
B\_purpose\_NHBO\_Car                       & -0.379 & 0.000        & 0.000        \\
B\_purpose\_NHBO\_Public\_Transport         & -0.715 & 0.000        & 0.000        \\
B\_start\_time\_linear\_Bike                & 0.017  & 0.000        & 0.015        \\
B\_start\_time\_linear\_Car                 & 0.027  & 0.000        & 0.000        \\
B\_start\_time\_linear\_Public\_Transport   & 0.010  & 0.000        & 0.016        \\
B\_traffic\_perc\_Car                       & -2.404 & 0.000        & 0.000        \\ \bottomrule
\end{longtable}
\end{center}

\begin{center}
\begin{longtable}{lrrr}
\caption{\centering Parameter estimates of the LPMC NL. Out of the 63 parameters, 10 are not significant at a 95\% confidence interval.}
\\ \toprule
\multicolumn{4}{c}{\textbf{LPMC - NL}}                                             \\ \midrule
                                            & Value  & Active bound & Rob. p-value \\ \midrule  \endhead 
ASC\_Bike                                   & -3.346 & 0.000        & 0.000        \\
ASC\_Car                                    & -2.439 & 0.000        & 0.000        \\
ASC\_Public\_Transport                      & -1.969 & 0.000        & 0.000        \\
B\_age\_Bike                                & -0.004 & 0.000        & 0.026        \\
B\_age\_Car                                 & 0.007  & 0.000        & 0.000        \\
B\_age\_Public\_Transport                   & 0.011  & 0.000        & 0.000        \\
B\_car\_ownership\_Bike                     & 0.062  & 0.000        & 0.351        \\
B\_car\_ownership\_Car                      & 0.628  & 0.000        & 0.000        \\
B\_car\_ownership\_Public\_Transport        & -0.037 & 0.000        & 0.369        \\
B\_con\_charge\_Car                         & -0.816 & 0.000        & 0.000        \\
B\_cost\_driving\_fuel\_Car                 & 0.000  & 1.000        & 1.000        \\
B\_cost\_transit\_Public\_Transport         & -0.077 & 0.000        & 0.000        \\
B\_day\_of\_week\_Bike                      & -0.023 & 0.000        & 0.155        \\
B\_day\_of\_week\_Car                       & 0.022  & 0.000        & 0.007        \\
B\_day\_of\_week\_Public\_Transport         & -0.033 & 0.000        & 0.000        \\
B\_distance\_Bike                           & -0.225 & 0.000        & 0.042        \\
B\_distance\_Car                            & -0.005 & 0.000        & 0.960        \\
B\_distance\_Public\_Transport              & 0.000  & 1.000        & 1.000        \\
B\_driving\_license\_Bike                   & 0.705  & 0.000        & 0.000        \\
B\_driving\_license\_Car                    & 0.484  & 0.000        & 0.000        \\
B\_driving\_license\_Public\_Transport      & -0.396 & 0.000        & 0.000        \\
B\_dur\_cycling\_Bike                       & -1.839 & 0.000        & 0.003        \\
B\_dur\_driving\_Car                        & -3.409 & 0.000        & 0.000        \\
B\_dur\_pt\_access\_Public\_Transport       & -3.410 & 0.000        & 0.000        \\
B\_dur\_pt\_bus\_Public\_Transport          & -1.445 & 0.000        & 0.000        \\
B\_dur\_pt\_int\_waiting\_Public\_Transport & -3.036 & 0.000        & 0.000        \\
B\_dur\_pt\_int\_walking\_Public\_Transport & -1.876 & 0.000        & 0.017        \\
B\_dur\_pt\_rail\_Public\_Transport         & -1.085 & 0.000        & 0.000        \\
B\_dur\_walking\_Walk                       & -8.171 & 0.000        & 0.000        \\
B\_female\_Bike                             & -0.831 & 0.000        & 0.000        \\
B\_female\_Car                              & 0.112  & 0.000        & 0.000        \\
B\_female\_Public\_Transport                & 0.156  & 0.000        & 0.000        \\
B\_fueltype\_Avrg\_Bike                     & -0.680 & 0.000        & 0.000        \\
B\_fueltype\_Avrg\_Car                      & -1.175 & 0.000        & 0.000        \\
B\_fueltype\_Avrg\_Public\_Transport        & -0.322 & 0.000        & 0.000        \\
B\_fueltype\_Diesel\_Bike                   & -0.817 & 0.000        & 0.000        \\
B\_fueltype\_Diesel\_Car                    & -0.251 & 0.000        & 0.000        \\
B\_fueltype\_Diesel\_Public\_Transport      & -0.391 & 0.000        & 0.000        \\
B\_fueltype\_Hybrid\_Bike                   & -0.985 & 0.000        & 0.000        \\
B\_fueltype\_Hybrid\_Car                    & -0.755 & 0.000        & 0.000        \\
B\_fueltype\_Hybrid\_Public\_Transport      & -0.941 & 0.000        & 0.000        \\
B\_fueltype\_Petrol\_Bike                   & -0.864 & 0.000        & 0.000        \\
B\_fueltype\_Petrol\_Car                    & -0.259 & 0.000        & 0.000        \\
B\_fueltype\_Petrol\_Public\_Transport      & -0.315 & 0.000        & 0.000        \\
B\_pt\_n\_interchanges\_Public\_Transport   & -0.088 & 0.000        & 0.093        \\
B\_purpose\_B\_Bike                         & -0.032 & 0.000        & 0.748        \\
B\_purpose\_B\_Car                          & -0.066 & 0.000        & 0.337        \\
B\_purpose\_B\_Public\_Transport            & -0.053 & 0.000        & 0.445        \\
B\_purpose\_HBE\_Bike                       & -1.036 & 0.000        & 0.000        \\
B\_purpose\_HBE\_Car                        & -0.665 & 0.000        & 0.000        \\
B\_purpose\_HBE\_Public\_Transport          & -0.271 & 0.000        & 0.000        \\
B\_purpose\_HBO\_Bike                       & -0.771 & 0.000        & 0.000        \\
B\_purpose\_HBO\_Car                        & -0.317 & 0.000        & 0.000        \\
B\_purpose\_HBO\_Public\_Transport          & -0.397 & 0.000        & 0.000        \\
B\_purpose\_HBW\_Bike                       & -0.272 & 0.000        & 0.000        \\
B\_purpose\_HBW\_Car                        & -0.976 & 0.000        & 0.000        \\
B\_purpose\_HBW\_Public\_Transport          & -0.570 & 0.000        & 0.000        \\
B\_purpose\_NHBO\_Bike                      & -1.235 & 0.000        & 0.000        \\
B\_purpose\_NHBO\_Car                       & -0.415 & 0.000        & 0.000        \\
B\_purpose\_NHBO\_Public\_Transport         & -0.678 & 0.000        & 0.000        \\
B\_start\_time\_linear\_Bike                & 0.016  & 0.000        & 0.016        \\
B\_start\_time\_linear\_Car                 & 0.024  & 0.000        & 0.000        \\
B\_start\_time\_linear\_Public\_Transport   & 0.012  & 0.000        & 0.002        \\
B\_traffic\_perc\_Car                       & -1.947 & 0.000        & 0.000        \\
MU\_m                                       & 1.391  & 0.000        & 0.000        \\ \bottomrule
\end{longtable}
\end{center}

\bibliographystyle{elsarticle-harv} 
\bibliography{cas-refs}





\end{document}